\documentclass[preprint, 12pt]{elsarticle}

\usepackage{geometry}
\usepackage{amssymb}
\usepackage{amsmath}
\usepackage{graphicx}
\usepackage{booktabs}
\usepackage{caption}
\usepackage{subcaption}
\usepackage{tabularx}
\usepackage{algorithm} 
\usepackage{algpseudocode}
\usepackage{url}
\usepackage{adjustbox}
\usepackage{rotating}   
\usepackage[table]{xcolor} 
\usepackage{array}      

\usepackage{soul}

\usepackage{tikz}
\usetikzlibrary{calc}
\usepackage{threeparttable}
\usepackage{rotating}
\usepackage{hyperref}
\usepackage{lineno}

\journal{Computers & Fluids}
\begin{document}


\begin{frontmatter}
    \title{XRePIT: A deep learning–computational fluid dynamics hybrid framework implemented in OpenFOAM for fast, robust, and scalable unsteady simulations.}

    \author[a]{Shilaj Baral} 
        \author[b]{Youngkyu Lee}
        \author[c]{Sangam Khanal}
        \author[a]{Joongoo Jeon\corref{cor1}}
        \cortext[cor1]{Corresponding author. Email address: jgjeon41@postech.ac.kr (Joongoo Jeon)}
        
        \affiliation[a]{
        organization={Division of Advanced Nuclear Engineering, Pohang University of Science and Technology},
        addressline={77 Cheongam-ro}, 
        city={Pohang-si},
        postcode={37673}, 
        state={Gyeongsangbuk-do},
        country={Republic of Korea}
        }
        
        \affiliation[b]{
        organization={Division of Applied Mathematics, Brown University},
        addressline={170 Hope Street}, 
        city={Providence},
        postcode={02906}, 
        state={Rhode Island},
        country={United States}
        }

        \affiliation[c]{
        organization={Graduate School of Integrated Energy-AI, Jeonbuk National University},
        addressline={567 Baekje-daero}, 
        city={Jeonju-si},
        postcode={54896}, 
        state={Jeollabuk-do},
        country={Republic of Korea}
        }
    
    \begin{abstract}
        Autoregressive neural surrogates offer computational acceleration for fluid dynamics but inherently suffer from error accumulation and non-physical drift during long-term rollouts. Although hybrid strategies combining surrogate models and physics-based solvers have been proposed, they are limited to manual implementations for low-dimensional benchmarks. In this study, we propose an OpenFOAM-based hybrid framework, XRePIT (eXtensible Residual-based Physics-Informed Transfer learning),  characterized by its fastness, robustness, and scalability. Unlike prior manual implementations (e.g., RePIT), XRePIT integrates a fully automated open-source workflow that manages the state transition between a neural surrogate and a traditional numerical solver (OpenFOAM) based on a monitored residual threshold. Using 3D buoyancy-driven flow as a testbed, we demonstrate that this residual-guided coupling enables stable long-term simulation—well beyond the stability horizon of standalone surrogates. Our results indicate that the hybrid loop achieves up to 2.91$\times$ wall-clock acceleration while maintaining relative $L^2$ errors within $\mathcal{O}(10^{-3})$. Furthermore, we benchmark the framework's extensibility by introducing a finite-volume-based Fourier neural operator (FVFNO), confirming that the stabilizing effect of the residual guardrail is agnostic to the underlying neural architecture. This study provides a deployable methodology for fast, robust, and automated hybrid simulation in 3D unsteady flow.
    \end{abstract}
    

\end{frontmatter}

\section{Introduction}
\label{sec:introduction}

Partial differential equations (PDEs) are the core mathematical language for modeling physical phenomena, describing how systems evolve in space and time. In fluid dynamics, the complexity of these equations in realistic settings necessitates computational fluid dynamics (CFD) to simulate their behavior. At the heart of CFD are three primary discretization strategies, each with a distinct mathematical foundation: the finite difference method (FDM), which uses Taylor expansions to approximate derivatives at discrete grid points~\cite{godunov1959fdm}; the finite volume method (FVM), which applies the divergence theorem to convert volume integrals into surface integrals~\cite{eymard2000fvm}; and the finite element method (FEM), which approximates weak solutions using nodal basis functions~\cite{dhatt2012fem}. A common thread among these methods is the need to discretize the spatial domain into a mesh. Capturing the intricacies of complex problems demands progressively finer meshes, which drastically increases computational cost~\cite{kim2005compCOST}. This burden is substantial: simulating mere seconds of physical time can require hours of dedicated CPU time~\cite{jeon2019identificationCPUtime1,tolias2018CPUtime2}. Furthermore, real-time control and design optimization in critical systems, such as small modular reactors (SMRs)~\cite{locatelli2014smr}, data-driven control with deep reinforcement learning (DRL)~\cite{jeon2024DRL,radaideh2021pof2,vignon2023pof3}, and data assimilation workflows are also hampered by the prohibitive cost of high-fidelity CFD simulations

To reduce this cost, machine learning (ML)—particularly neural networks (NNs)—has emerged as a promising accelerator. NNs can approximate input–output mappings given sufficient data and tuning~\cite{hornik1989multilayer}. Recent efforts apply ML to speed up CFD workflows~\cite{vinuesa2020SustainableDev,MLCFD3,kochkov2021machine,MLCFD2,vinuesa2022enhancing,SINHA2025RBF,RAUT2025GNN,NGUYEN2025DGNet}. Unlike traditional solvers that rely on sequential updates, neural models exploit GPU parallelism to produce rapid predictions.

In the realm of fluid mechanics, convolutional neural networks (CNNs) have been used to model steady-state flows by capturing spatial correlations~\cite{guo2016CNNForFluids, krizhevsky2012imagenet, CNN1ComputerAndFluids}, while recurrent neural networks (RNNs) model temporal evolution in unsteady flows~\cite{mohan2018RNNforFluids}. Advanced architectures like ConvLSTM, U-Net, and AutoEncoders have also been benchmarked in recent studies~\cite{khanal2025sangam} for transient flows. Beyond classical deep learning models, modern generative approaches such as diffusion models can reconstruct high-fidelity flow fields from coarse inputs~\cite{shu2023DiffusionModel}, and vision transformers (ViTs) are gaining popularity for their ability to capture both local and global dynamics~\cite{kang2023visionTransformer}. Some models explicitly embed physical laws into their design. For example, physics-informed neural networks (PINNs) include governing equations in the loss function~\cite{raissi2019PINNs, gao2021phygeonet, PINN1ComputerAndFluids, shin2025node, ZHAO2025PINNReview,PINN2ComputerAndFluids,zhao2024pinnpof4,PINN3ComputerAndFluids}, while others like the finite volume method network (FVMN)~\cite{jeon2022FVMN} structurally encode computational laws into the architecture itself. A significant paradigm shift has also occurred with the development of neural operators (NOs), such as the Fourier neural operator (FNO)~\cite{li2020FNO} and deep operator network (DeepONet)~\cite{lu2019deeponet}, which learn mappings between entire function spaces rather than pointwise values ~\cite{YE2024LNO, WANG2025En-DeepONet, NO1ComputerAndFluids, YANG2025MF-DeepONet,NO2ComputerAndFluids}. Table~\ref{tab:ai_cfd} lists representative studies applying ML to accelerate fluid simulations and briefly notes their application areas (with steady/laminar context where relevant).

\begin{table*}[htbp]
  \centering
  \caption{Selected AI-based strategies for accelerating CFD simulations.
  Abbreviations: NS = Navier--Stokes; RBC = Rayleigh--B\'enard convection; 
  LI = learned interpolation; GCN = graph convolutional network; 
  NO = neural operator; PPE = pressure Poisson equation.}
  \label{tab:ai_cfd}
  \setlength{\tabcolsep}{3.5pt}
  \renewcommand{\arraystretch}{1.15}
  \footnotesize
  \begin{tabular}{@{} l l l p{0.45\textwidth} @{}}
    \toprule
    \textbf{Year} & \textbf{Author} & \textbf{Technique} 
    & \textbf{Application} \\
    \midrule
    \multicolumn{4}{@{}l}{\textit{Standalone AI surrogates}} \\[2pt]
    2016 & Guo et al.~\cite{guo2016CNNForFluids}
         & CNN
         & Steady incompressible NS \\
    2018 & Mohan \& Gaitonde~\cite{mohan2018RNNforFluids}
         & LSTM
         & Turbulent flow ROM (cylinder wake) \\
    2019 & Raissi et al.~\cite{raissi2019PINNs}
         & PINN
         & NS inverse problem (cylinder wake) \\
    2020 & Li et al.~\cite{li2020FNO}
         & FNO
         & Burgers, Darcy flow, NS (vorticity) \\
    2021 & Lu et al.~\cite{lu2019deeponet}
         & DeepONet
         & Advection, diffusion--reaction \\
    2021 & Gao et al.~\cite{gao2021phygeonet}
         & PhyGeoNet
         & Steady convection--diffusion (irregular domains) \\
    2022 & Jeon et al.~\cite{jeon2022FVMN}
         & FVMN
         & Unsteady reacting \& non-reacting NS \\
    2023 & Shu et al.~\cite{shu2023DiffusionModel}
         & Diffusion model
         & Super-resolution of Kolmogorov flow \\
    2023 & Kang et al.~\cite{kang2023visionTransformer}
         & ViT + U-Net
         & Steady flow approximation \\
    2024 & Kontolati et al.~\cite{kontolati2024L-DeepONet}
         & L-DeepONet
         & RBC, shallow water equations \\
    2024 & Oommen et al.~\cite{oommen2024diffusion2}
         & NO + Diffusion
         & Buoyancy-driven NS, Kolmogorov flow \\
    2025 & Khanal et al.~\cite{khanal2025sangam}
         & CNN benchmark
         & Unsteady buoyancy-driven NS \\
    2025 & Wang et al.~\cite{wang2025flowmatching1}
         & Flow matching
         & Burgers, Kolmogorov flow \\
    2025 & Parikh et al.~\cite{parikh2025flowmatching2}
         & Flow matching
         & Near-wall turbulence (channel flow) \\
    \addlinespace[6pt]
    \multicolumn{4}{@{}l}{\textit{Hybrid AI--CFD methods}} \\[2pt]
    2020 & Belbute-Peres et al.~\cite{belbute2020combining}
         & GCN + diff.\ solver
         & Steady airfoil flow \\
    2021 & Kochkov et al.~\cite{kochkov2021machine}
         & CNN (LI) inside FVM
         & 2D turbulence \\
    2024 & Zhang et al.~\cite{HINTS}
         & NO + relaxation
         & Helmholtz, advection--diffusion \\
    2024 & Oommen et al.~\cite{oommen2024rethinking}
         & NO + DNS
         & Phase-field \\
    2024 & Sousa et al.~\cite{hybrid2}
         & PCA + MLP for PPE
         & Vortex shedding \\
    2024 & Jeon et al.~\cite{jeon2024residual}
         & FVMN + CFD (RePIT)
         & Buoyancy-driven NS \\
    2025 & Lee et al.~\cite{lee2025fastmetasolvers3dcomplexshape}
         & DeepONet + Krylov/relaxation
         & 3D Helmholtz scattering (complex geometries)\\
    \bottomrule
  \end{tabular}
\end{table*}

Despite rapid progress, important limitations remain. Many models target laminar flows or simplified physics~\cite{guo2016CNNForFluids}, and several struggle to maintain stability over long rollouts~\cite{khanal2025sangam}. PINNs can be computationally expensive and may underperform with dynamic boundary conditions or stiff PDEs~\cite{PINNsFailure}. A common ML design for CFD forecasting is auto-regressive prediction, where the model reuses its past outputs as inputs. Given a learned map $\mathcal{N}_\theta$, the update at time $t{+}1$ is
\begin{equation}
  Z^{t+1} = \mathcal{N}_\theta\!\left(Z^{t}\right),
\end{equation}
where $Z^t$ collects field variables (e.g., velocity $\boldsymbol{u}$, pressure $p$, temperature $T$). While simple and fast, this approach can accumulate error over time and drift toward non-physical states.

Neural operators offer an alternative, but generalization often depends on the training distribution. For example, including time in the trunk network of DeepONet can reduce error accumulation by learning a global mapping,
\begin{equation}
  \mathcal{G}:(f,t)\mapsto Z(t),
\end{equation}
where $f$ encodes initial/boundary conditions and $t$ is query time. However, such models are typically tied to the regimes seen during training, making true extrapolation to unseen time windows difficult.

To address the limits of purely data-driven approaches, hybrid frameworks combining AI with traditional CFD have gained traction. These generally fall into two categories: (i) \emph{iteration-coupled} methods~\cite{HINTS,lee2025fastmetasolvers3dcomplexshape,hybrid2}, in which ML components accelerate sub-steps inside each CFD iteration, and (ii) \emph{timestep-coupled} methods~\cite{jeon2024residual,oommen2024rethinking}, in which an ML model advances the solution across timesteps and falls back to CFD as needed. Iteration-coupled designs often require deep modifications of solver internals, complicating integration with widely used platforms like OpenFOAM. In contrast, timestep-coupled approaches preserve solver modularity by interfacing externally. We therefore adopt the timestep-coupled path, which naturally suits transient problems and allows interchangeable auto-regressive accelerators, making it ideal for systematic benchmarking.

Within this modular, timestep-coupled paradigm, RePIT was introduced as a proof-of-concept hybrid loop that alternates fast autoregressive prediction with residual-triggered CFD correction~\cite{jeon2024residual}. However, its original realization relied on manual coupling and a limited 2D evaluation, which made it difficult to run systematic, reproducible studies across residual thresholds, online update budgets, architectures, and boundary conditions.

We address these gaps with XRePIT (e\textbf{X}tensible RePIT): a fully automated, open-source hybrid CFD--ML framework built around OpenFOAM~\cite{jasak2007openfoam}. XRePIT operationalizes timestep-coupled hybridization into a reusable pipeline that executes ML rollouts, monitors a physics residual online (here, a mass-conservation residual), triggers on-demand CFD correction, and performs lightweight online transfer learning to keep the surrogate synchronized with evolving regimes. While the switching signal is intentionally simple and effective for the studied cases, we treat this single-residual design as a strong baseline (as verified in ~\cite{jeon2024residual}) and explicitly note that extensions to multi-residual switching(e.g., momentum/energy) are a natural next step for more complex unsteady regimes.

Building on the foundations and limitations described above, this paper makes the following contributions:
\begin{enumerate}
  \item \textbf{Automated residual-guided hybrid solver (XRePIT).} We deliver a fully automated timestep-coupled CFD--ML framework for OpenFOAM that converts residual-guided hybrid rollouts from a manual proof-of-concept into a reproducible and extensible workflow (including switching, correction, and online updates).

  \item \textbf{Systematic accuracy--speed characterization of residual thresholding and online updates.} We quantify how the residual threshold and transfer-learning budget govern the stability--acceleration trade-off, showing that small online updates can stabilize long rollouts, while overly aggressive updates provide diminishing returns in acceleration for the studied buoyancy-driven regimes.

  \item \textbf{Boundary-condition adaptation via online transfer learning (within fixed geometry).} We demonstrate that a surrogate trained under one boundary condition can be reused under other boundary conditions of the same configuration through lightweight online adaptation, preserving the flow regime while maintaining bounded error growth.

  \item \textbf{Architecture interchangeability inside the same hybrid loop.} To show that the framework is not tied to a single surrogate family, we introduce a finite-volume-based Fourier neural operator (FVFNO) and benchmark it against the original finite-volume-based MLP (FVMN) within the identical residual-guided correction mechanism~\cite{jeon2022FVMN}.

  \item \textbf{3D extension of residual-guided timestep-coupled hybridization.} We extend the full pipeline to a three-dimensional buoyancy-driven case to evaluate scalability under increased state dimension, demonstrating that residual-guided correction remains an effective stabilizer and can yield practical acceleration in 3D.
\end{enumerate}

Together, these contributions position XRePIT as a systematized and automated testbed for residual-guided timestep-coupled hybrids---enabling controlled comparisons across thresholds, update schedules, architectures, and boundary conditions, and providing evidence that hybrid correction can deliver stable long-horizon acceleration in settings where pure autoregressive ML rollouts drift.

The remainder of the paper is organized as follows. Section~\ref{sec:foundational_works} summarizes the baseline RePIT formulation and data-efficient FVMN architecture. Section~\ref{sec:methodology} describes the numerical setup and the automated residual-guided hybrid loop. Section~\ref{sec:results} presents the 2D stability analyses, boundary-condition transfer experiments, architecture benchmarking, and the 3D scalability study. Section~\ref{sec:discussion} concludes with key findings, limitations, and extensions.

\section{Foundational Works}
    \label{sec:foundational_works}

    This section briefly lays the foundation for the timestep-coupled RePIT concept and a grid-based neural network that was used in a previous study, a fine-tuned version of which is employed in the present study.
    
    \subsection{Vanilla RePIT}
    \label{subsec:vanilla_repit}
    This approach started with running a traditional CFD simulation in OpenFOAM for an initial sequence of $N$ time steps. This behaved like any typical transient solver and produced ground-truth flow fields, such as $\mathbf{u}^0,\mathbf{u}^1,\dots,\mathbf{u}^N$, based purely on physics. These field values were then manually extracted and saved into a CSV file, organized by time step, which was used as the training dataset for a grid-based neural network (FVMN). After manual pre-processing, the model was trained using this data. 
    
    During inference, the quality of ML predictions was monitored using a scaled residual metric ($R_{\text{rel}}$), derived from the mass conservation equation. As formulated in Equation~\eqref{eq:relative_residual_limit}, the residual was computed at each predicted timestep and normalized by a fixed reference value $R_{\text{mass}}^{t_0}$, obtained from the initial training timestep. The residual itself was calculated using a central difference approximation shown in Equation~\eqref{eq:R_mass}, capturing the imbalance in continuity across all grid cells with velocities $u, v$ defined on an $n_x \times n_y$ mesh. Unlike unscaled residuals, which lack an absolute threshold for interpretation, scaling against a known baseline provided a consistent convergence indicator. And if the scaled residual exceeded a threshold (set to 5), the ML inference was stopped. For example, if the threshold is crossed at $\mathbf{u}^{N+M}$, prediction halts there, where $M$ is the ML predicted time steps. Before reinserting these fields, boundary consistency is ensured by directly attaching boundary values from the ground-truth CFD data. For example, if the grid is 200×200, the ML model predicts values for the 198×198 internal points, while the boundary layers are filled using known values from the CFD snapshots. This reliance on ground truth data is also one of the major limitations of this strategy ~\cite{jeon2024residual}.
    
    \begin{equation}
        \label{eq:relative_residual_limit}
            R_{\text{rel}} = \frac{R_{mass}^{t}}{R_{mass}^{t_0}}
    \end{equation}
    
    \begin{equation}
    \label{eq:R_mass}
        R_{mass}^{t} = \frac{1}{n_x n_y}\sum_{i=1}^{n_x-1}\sum_{j=1}^{n_y-1}\bigg(\frac{u_{i+1,j}-u_{i-1,j}}{2\Delta x}+\frac{v_{i,j+1}-v_{i,j-1}}{2\Delta y}\bigg)^2
    \end{equation}
    
\begin{figure}[htpb]
        \centering
        \includegraphics[width=1.0\textwidth]{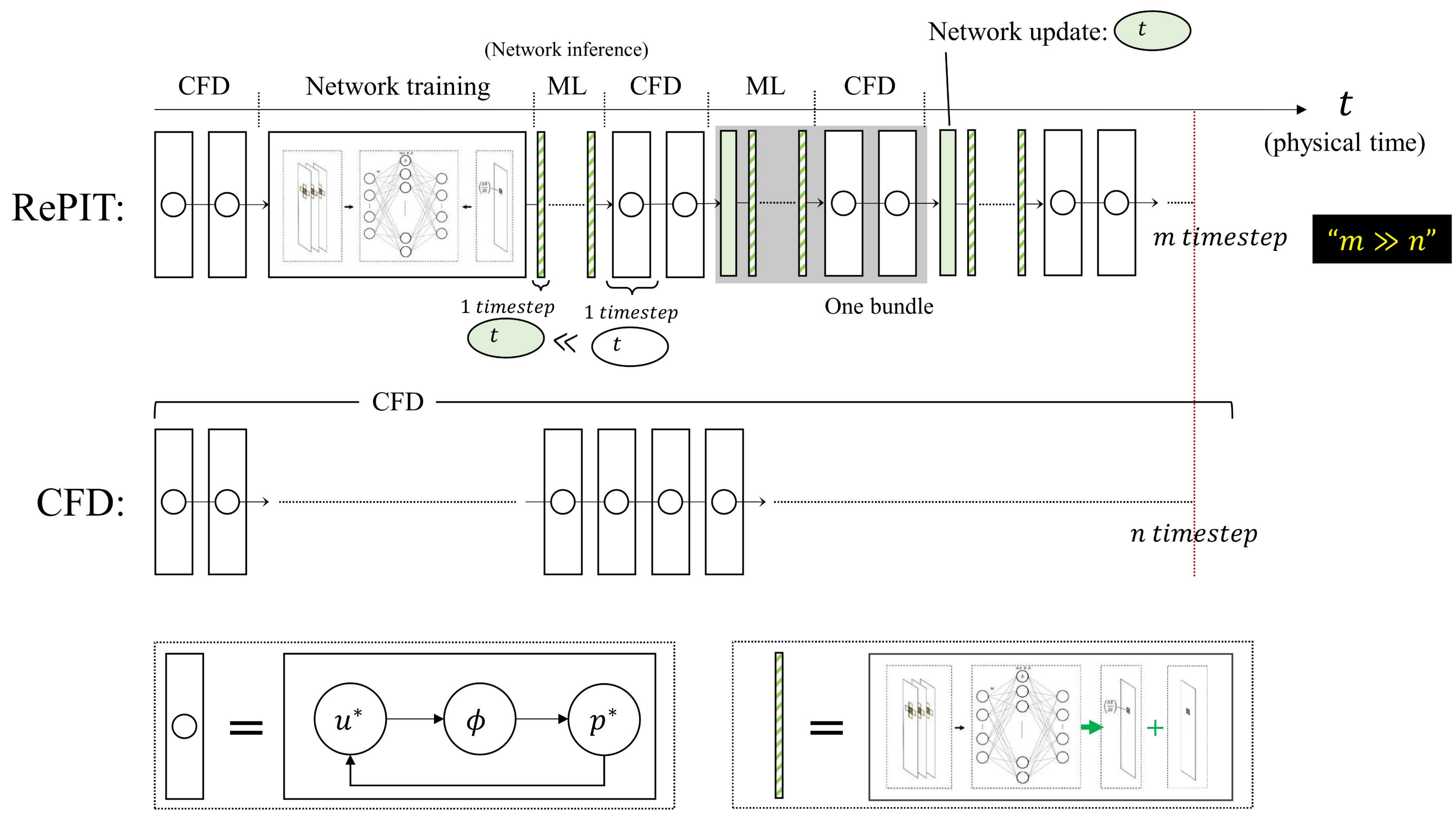} 
        \caption{\textbf{Schematic overview of the vanilla RePIT workflow.}}
        \label{fig:vanilla_repit}
    \end{figure}

The predicted field variables were manually copied and pasted back into OpenFOAM's native data structure. The time it takes for ML to predict a single timestep of simulation is much less compared to the solver calculation. As shown in Fig. ~\ref{fig:vanilla_repit}, this increased time for the CFD solver is due to the iterative correction of the velocity and pressure fields, while ML inference is non-iterative and can be parallelized. After the conversion, the new simulation run produced higher-fidelity results, which were then used to fine-tune the same neural network via transfer learning. Re-training was performed for 10 epochs on each updated dataset. This entire process is repeated approximately 30 times to complete 1000 timesteps. And as a whole, this network update, solver calculation, and neural network prediction, along with pre- and post-processing routines, can be thought of as one bundle of this hybrid workflow. For the same physical time, RePIT can compute a much larger number of timesteps compared to conventional CFD.

However, this ``vanilla'' RePIT implementation had several practical limitations:
\begin{enumerate}
    \item \emph{Boundary-condition teacher forcing:} relying on CFD ground truth to overwrite all boundary values prevented the RePIT from acting as a truly stand-alone hybrid solver and limited its applicability to cases where full CFD data were always available.

    \item \emph{Manual data exchange:} the coupling between OpenFOAM and the ML model relied on ad hoc file manipulations rather than a systematic, reusable data exchange mechanism.
    
    \item \emph{Need for automation:} The RePIT study was largely manual, obscuring its application and analysis to a wide range of case studies and making it harder to reproduce at scale.
    
    \item \emph{Limited validation of scalability and generalization:} the method was tested only on a single 2D natural-convection configuration, without exploring different dimensions, neural architectures, or even boundary conditions. Hence, it could not provide enough evidence for the methods scalability and generalizability.
\end{enumerate}

These constraints motivated the development of XRePIT: a fully automated, extensible hybrid ML-CFD framework that removes manual steps, enforces boundary conditions within the ML pipeline itself, formalizes the data exchange, and systematically evaluates scalability and generalization across boundary cases, network architectures, and higher-dimensional problems.
    
\subsection{Finite volume method network (FVMN)}
\label{subsec:fvmn}
   The vanilla RePIT strategy was tied with the FVMN neural architecture ~\cite{jeon2022FVMN}. It was specifically chosen for its remarkable data efficiency, making it a critical feature that reinforces a hybrid strategy, as it demanded fewer CFD solver calls for both initial training and subsequent online adaptation, thereby maximizing the potential for acceleration. The data-efficiency is brought about by the three core principles specified below:

\begin{enumerate}
    \item \textbf{Tiered stencil input:} Inspired by the finite volume method, the input for each cell is a feature vector comprising the cell's own value and the values of its immediate neighbors. This provides the network with local spatial context, analogous to a numerical discretization stencil.
    \item \textbf{Variable specific subnetworks:} Instead of a single monolithic model, the architecture uses separate, independent subnetworks for each physical variable (e.g., velocity components and temperature). The final prediction is an aggregation of the outputs, and the network is trained on a combined loss function.
    \item \textbf{Derivative output:} The network is trained to predict the temporal derivative of the field variables ($\Delta Z / \Delta t$), rather than the absolute state at the next timestep. The final prediction, $Z^{t+1}$, is obtained by adding this predicted change to the current state, $Z^t$.
\end{enumerate}

\section{Methods}\label{methodology}
\label{sec:methodology}

\subsection{Numerical simulation setup}
\label{sec:numerical_setup}

\subsubsection{Governing physics and solver}
The study investigates a natural convection flow where heat transfer is the primary driver. To capture the underlying physics, we employed the \texttt{buoyantFoam} solver from OpenFOAM v13. This is a transient solver designed for buoyancy-driven flows of compressible fluids and has been validated against experimental results for similar thermal problems \cite{nielsen1976solverValidation, kit2023solverValidation}. The working fluid was modeled as air, treated as a perfect gas where density is computed using the ideal gas law. Although this is a compressible formulation, the pressure variation across the domain was negligible in our simulations. Consequently, the fluid density was primarily a function of temperature, resulting in flow behavior that closely resembles that of an incompressible flow under the Boussinesq approximation \cite{ferziger2002computational(boussinesq)}.

The \texttt{buoyantFoam} solver addresses non-isothermal, compressible flow by solving the conservation equations for mass (Eq.~\ref{eq:continuity_equation}), momentum (Eq.~\ref{eq:momentum_equation}), and energy (Eq.~\ref{eq:energy_equation}). We note that Eq.~\ref{eq:energy_equation} is 
written in terms of enthalpy $h$ rather than internal energy $e$. Since enthalpy is 
defined as $h = e + p/\rho$, the total energy per unit volume satisfies 
$\rho e = \rho h - p$, and taking the time derivative yields 
$\partial(\rho e)/\partial t = \partial(\rho h)/\partial t - \partial p/\partial t$. 
Consequently, when the standard total energy conservation equation is recast in terms 
of $h$, the $-\partial p/\partial t$ term appears explicitly on the left-hand side. In these equations, $t$ is time, $\rho$ is the density field, and $\mathbf{u}$ is the velocity field. For the momentum equation, $p$ represents the static pressure field, $\mathbf{g}$ is the gravitational acceleration, and $\mu_{eff}$ is the effective dynamic viscosity, which is the sum of the molecular and turbulent viscosities. 
In the energy equation, $K$ is the kinetic energy per unit mass ($K \equiv |\mathbf{u}|^2/2$), and $\alpha_{eff}$ is the effective thermal diffusivity, which combines laminar and turbulent thermal diffusivities.

\begin{equation}
        \label{eq:continuity_equation}
            \frac{\partial \rho}{\partial t} + \nabla \cdot (\rho \mathbf{u})=0
    \end{equation}

    \begin{equation}
        \label{eq:momentum_equation}
        \begin{split}
            \frac{\partial (\rho \mathbf{u})}{\partial t}+\nabla \cdot \rho (\mathbf{u}\otimes\mathbf{u}) =-\nabla p + \rho g + \nabla \cdot \bigg(\mu_{eff} (\nabla \textbf{u} + \nabla \textbf{u}^T)\bigg) - \nabla \bigg(\frac{2}{3}\mu_{eff} (\nabla \cdot \textbf{u})\bigg)
        \end{split}
    \end{equation}

    \begin{equation}
        \label{eq:energy_equation}
            \frac{\partial (\rho h)}{\partial t}+\nabla \cdot (\rho \mathbf{u} h)+\frac{\partial (\rho K)}{\partial t} + \nabla \cdot (\rho \mathbf{u} K) - \frac{\partial p}{\partial t} = \nabla \cdot (\alpha_{efff}\nabla h) + \rho \textbf{u}\cdot g
    \end{equation}

\subsubsection{Discretization and linear solvers}
The pressure-velocity coupling was managed by the PIMPLE algorithm, a hybrid of the PISO and SIMPLE algorithms ~\cite{barton1998SIMPLE}, configured with two inner correction loops and one outer correction loop per time step. Temporal discretization was handled using a first-order implicit Euler scheme. For spatial discretization, a second-order accurate finite volume method was employed with the following schemes: a \texttt{Gauss linear} scheme for gradient terms, a \texttt{Gauss upwind} scheme for convective terms to ensure stability, and a \texttt{Gauss linear corrected} scheme for Laplacian terms ~\cite{eymard2000fvm}.

The resulting linear systems were solved using iterative methods: the pressure equation (\texttt{p\_rgh}) was solved using a Preconditioned Conjugate Gradient (PCG) ~\cite{lee1999PCG} solver with a Diagonal Incomplete-Cholesky (DIC) preconditioner ~\cite{meurant2001DIC}, while the momentum and energy equations were solved using a Preconditioned Bi-Conjugate Gradient Stabilized (PBiCGStab) solver ~\cite{mikic1985PBICGStab} with a Diagonal Incomplete-LU (DILU) preconditioner ~\cite{zhang2001DILU}.

\subsubsection{Computational domain and case definitions}
\label{subsubsec:domain_and_case}
Each snapshot in time contains full-field data for temperature $T(x,y,t)$, and velocity $\mathbf{u}(x,y,t)$ on a $200\times200$ grid. This setup remains consistent with prior benchmark study ~\cite{jeon2024residual}, allowing for reliable cross-validation and repeatability. Our preliminary investigation using the same dataset reveals that even advanced CNN-based architectures, when trained on 800 timesteps of natural convection data, can reliably predict only the first ten steps into the future ~\cite{khanal2025sangam}. This highlights the inherent difficulty of learning the evolving flow dynamics, particularly in the boundary regions and also when the network is expected to extrapolate far beyond the training horizon. 

To evaluate the generalization of the hybrid method, three distinct cases with different thermal boundary conditions were studied. For all cases, the top and bottom walls were adiabatic (zero-gradient, Neumann condition), and a no-slip condition was applied to all walls. The cases differ by the Dirichlet conditions on the vertical walls:
\begin{itemize}
    \item \textbf{Case 1 (Baseline):} Hot wall at 307.75 K, cold wall at 288.15 K.
    \item \textbf{Case 2:} Hot wall at 317.75 K, cold wall at 278.15 K.
    \item \textbf{Case 3:} Hot wall at 327.75 K, cold wall at 268.15 K.
\end{itemize}
The flow regime for the baseline case (Case 1) is characterized by a Rayleigh number of \textbf{$Ra = 1.85 \times 10^9$} and a Prandtl number of \textbf{$Pr = 0.705$}.

The extension to 3D is in the spatial domain of  $0.75\text{m} \times 0.75\text{m} \times 1.5\text{m}$, similar to the experimental setup defined in ~\cite{ampofo2003experimental}. This 3D case is not a simple extrusion; it introduces a spanwise (z-direction) degree of freedom, allowing for the formation of more intricate 3D flow structures that are fundamentally absent in the 2D approximation.

\begin{figure}[t]
    \centering
    \includegraphics[width=1.0\textwidth]{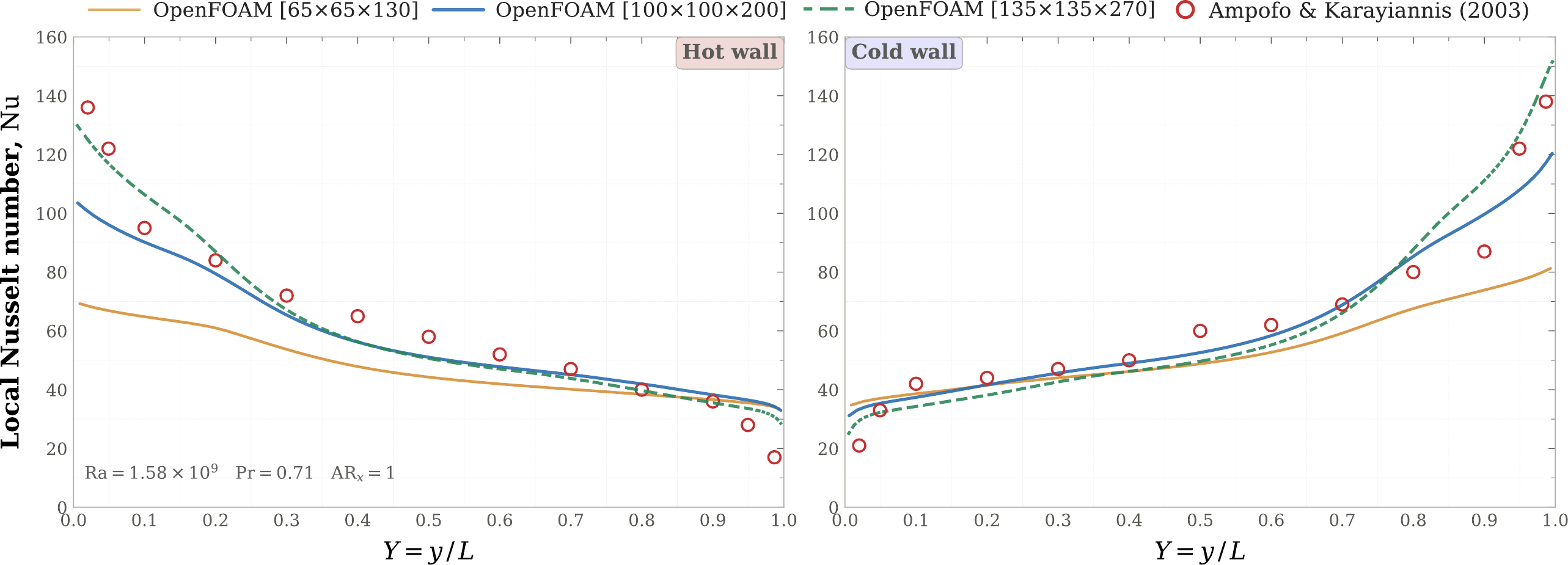}
    \caption{Grid convergence study for the 3D natural convection case. Local 
    Nusselt number along the hot and cold wall for three mesh resolutions compared against the experimental data of~\cite{ampofo2003experimental}. The medium grid 
    ($100 \times 100 \times 200$) provides sufficient accuracy and is adopted for 
    all 3D simulations in this work.}
    \label{fig:grid_convergence}
\end{figure}

 To select an appropriate spatial resolution, a grid convergence study was 
conducted using three uniform meshes: a coarse grid ($65 \times 65 \times 130$), 
a medium grid ($100 \times 100 \times 200$), and a fine grid 
($135 \times 135 \times 270$). The steady-state local Nusselt number distribution 
along the hot and cold walls were compared against the experimental measurements 
of~\cite{ampofo2003experimental}, as shown in Fig.~\ref{fig:grid_convergence}. 
The coarse mesh exhibits visible deviations, particularly in the boundary layer 
regions near the top and bottom of the cavity, whereas the medium and fine meshes 
both capture the overall Nusselt number profile with reasonable agreement to the 
experimental data. Since the difference between the medium and fine grids is 
marginal while the computational cost increases substantially, the medium grid 
($100 \times 100 \times 200$) was selected for all subsequent 3D simulations. A 
detailed summary of the grid convergence metrics is provided in 
~\ref{sec:grid_convergence}.
For this 3D case, the new front and back walls were also defined as adiabatic 
and no-slip, consistent with the top and bottom walls, and we refer to it as Case 4 throughout the manuscript.

\subsection{Modular architecture of hybrid workflow}\label{method:hybrid_workflow}

\begin{figure*}[htbp]
    \centering
    \includegraphics[width=0.8\textwidth]{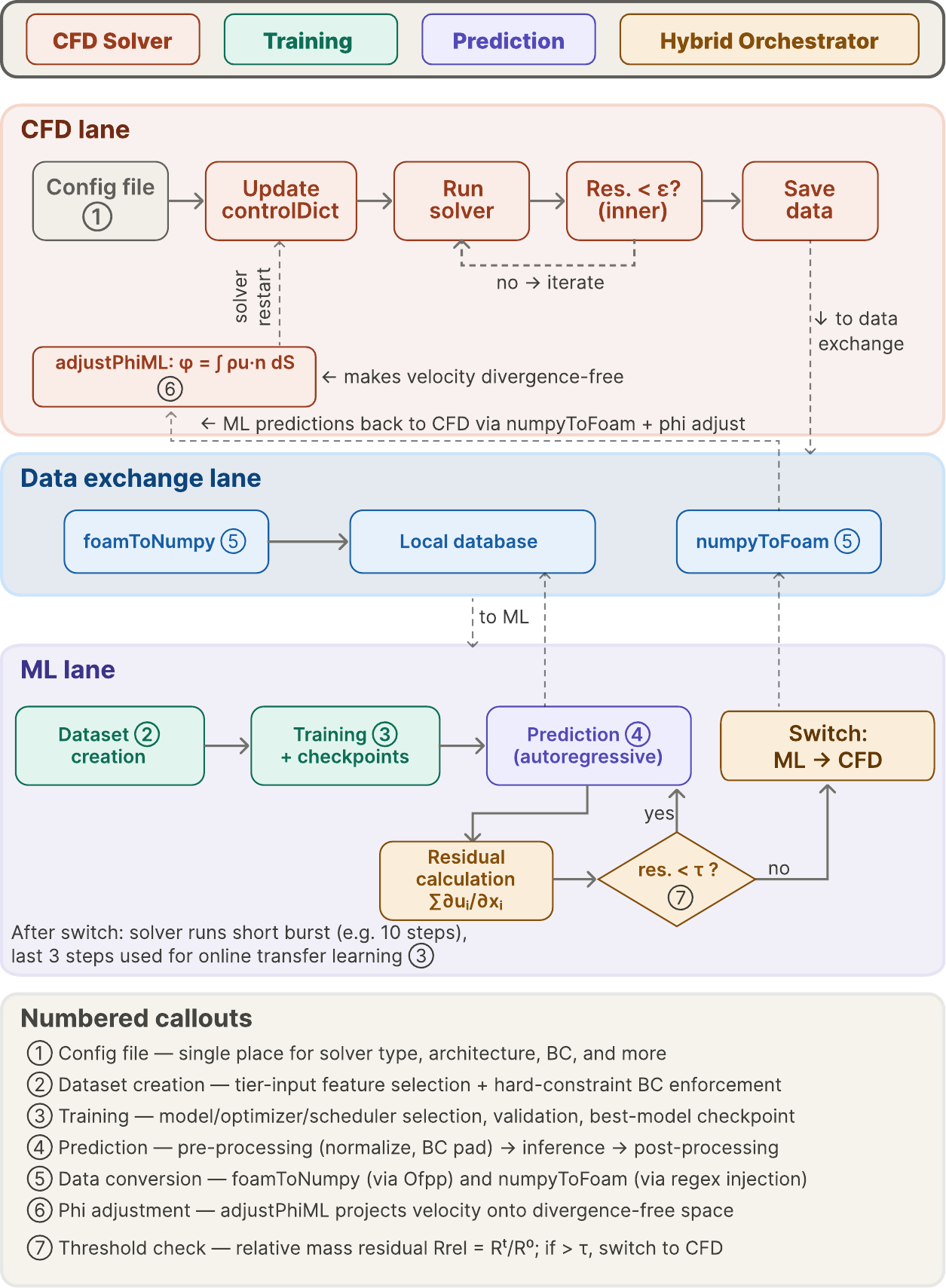}
    \caption{\textbf{Schematic of the automated, timestep-coupled hybrid workflow.} The loop alternates between rapid, auto-regressive prediction by the ML surrogate and on-demand, single-step correction by the CFD solver, triggered by a residual-guided switching logic.}
    \label{fig:methods_fig1}
\end{figure*}

The research was conducted using XRePIT, a novel fully automated framework designed to orchestrate hybrid ML-CFD simulations. The framework is built on a modular, Python-based ecosystem that separates the core responsibilities of simulation control, machine learning, and solver interaction, as illustrated in Fig. \ref{fig:methods_fig1}. The entire process is managed from a single configuration file where users define all simulation, model, and hybrid control parameters.


\subsubsection{The hybrid orchestrator}

At the core of the framework is the hybrid orchestrator. This is the master control script that executes the main hybrid loop and manages the adaptive switching logic. It initiates the simulation by calling the CFD solver for an initial data generation phase. Subsequently, it directs the Predictor to perform auto-regressive rollouts, continuously monitoring a physics-based residual calculated from the ML-predicted fields. When the residual exceeds a pre-defined threshold, the Orchestrator halts the ML rollout and triggers a two-step correction: it first calls the Solver Interface to obtain a high-fidelity correction from OpenFOAM and then instructs the Trainer to perform online transfer learning using this new data. As shown in Algorithm ~\ref{alg:repit}, the use of the residual value to determine the convergence criteria is widely-adopted in CFD solvers, whereas in the AI part, we use the same metric to determine the divergence criteria. It can be seen from steps 5 to 9 in the algorithm that the internal iteration is repeated unless the residual value \textbf{falls} below a certain threshold. The iteration-coupled hybrid method uses neural networks to reduce this iteration. Whereas timestep-coupled hybrid methods like XRePIT use the neural network to predict the whole new timestep data until it \textbf{crosses} a certain threshold (as shown from steps 15 to 20). The training, prediction, data-conversion, and phi-adjustment logic are discussed in the subsequent sections.

 To support long-running engineering simulations, the framework provides continuous checkpointing across all modules. The Trainer saves the surrogate model's weights, optimizer state, and learning rate scheduler to a local database after each validation improvement, alongside the data normalization statistics (means and standard deviations) that are written during dataset creation. The hybrid orchestrator independently maintains a running log of simulation metrics, including the physical residual history. In the event of an intentional halt or unexpected hardware failure, a user can seamlessly resume the hybrid simulation by updating the central configuration file with the latest saved OpenFOAM timestep. Upon restart, the framework reloads the most recent surrogate checkpoint and its corresponding normalization statistics, and re-enters the hybrid loop from the last CFD-corrected state—avoiding any redundant computation or data loss.

\begin{algorithm}[htbp]
    \caption{The main hybrid loop (XRePIT)}
    \label{alg:repit}
    \begin{algorithmic}[1]
        \State \textbf{Input:} initial neural network parameters $\theta_0$, initial flow fields, $\varepsilon_{\text{th}}$, solver parameters
        \While {$t \in [0, T]$} \Comment{Time stepping loop}
            \For {$k=t,t+t_{\text{CFD}}$} \Comment{CFD Section}
                \State Discretize momentum equation using finite volume method:
                \[\int_V \frac{\partial \mathbf{u}}{\partial t} dV + \int_S (\mathbf{u} \otimes \mathbf{u}) \cdot \mathbf{n} dS - \int_S (\nu \nabla \mathbf{u}) \cdot \mathbf{n} dS = -\int_V \nabla p dV\]
                \While{$\varepsilon > \varepsilon_{\text{th}}$}
                    \State Correct velocity equation: 
                    $\mathbf{u}^{*} \leftarrow \mathbf{u} = \frac{H}{A} - \frac{1}{A} \nabla p$
                    \State Solve pressure equation using the continuity equation:
                    \[
                    p^{\text{new}} \leftarrow \nabla \cdot \left( \frac{1}{A} \nabla p \right) = \nabla \cdot \left( \frac{H}{A} \right)
                    \]
                    \State Update continuity error $\varepsilon$
                \EndWhile
                \State Apply momentum corrector:
                $\mathbf{u}^{\text{new}} \leftarrow \frac{H}{A} - \frac{1}{A}\nabla p$
                \State $t = t+\Delta t$
            \EndFor
            \State Convert OpenFOAM type data to numpy
            \State Optimize surrogate loss $L$ w.r.t. $\theta$ \Comment{ML Section}
            \[
            \theta^*= \arg\min_{\theta} \sum ([\mathbf{\hat{u}}^{\text{new}}, \hat{p}^{\text{new}}]-[\mathbf{u}^{\text{new}},p^{\text{new}}])^2
            \]
            \While{$\varepsilon \leq \varepsilon_{\text{th}}\ \textbf{and}\ t_i \leq T$} 
                \State $\mathbf{u},p \leftarrow \mathbf{u}^{\text{new}}, p^{\text{new}}$
                \State $\mathbf{u}^{\text{new}}, p^{\text{new}} = \mathcal{N}(\mathbf{u},p;\theta^*)$
                \State $t = t+\Delta t$
                \State Update continuity error $\varepsilon$
            \EndWhile
            \State Convert numpy data to OpenFOAM type using \texttt{numpyToFoam} utility
            \State Run \texttt{adjustPhi} utility to correct the flux field $\phi$
            \State Repeat until $t==T$
        \EndWhile
    \State \textbf{Output:} Calculated field values up to timestep $T$ using hybrid-computation
    \end{algorithmic} 
\end{algorithm}

\subsubsection{Machine learning modules}

The machine learning responsibilities are handled by two specialized modules:

\textbf{The Trainer:} This module encapsulates all aspects of model training and adaptation. It is responsible for selecting the specified model architecture, optimizer, and learning rate scheduler from a unified configuration. The Trainer first conducts the initial training of the surrogate model from scratch on the dataset generated by OpenFOAM. It also manages the online transfer learning process when triggered by the Orchestrator after a CFD correction. It fine-tunes the model's weights on a small buffer of new, high-fidelity data. The Trainer continuously tracks validation loss and saves the best-performing model checkpoint, which is then used for the subsequent prediction phase.
 
\textbf{The Predictor:} This lightweight module is dedicated solely to high-speed inference and is responsible for executing the auto-regressive ML rollouts. When called by the Orchestrator, the Predictor enters a loop where it performs the following sequence at each timestep: (i) it pre-processes the input data, including normalization and the enforcement of boundary conditions; (ii) it passes the prepared data to the trained model for inference; (iii) it post-processes the model's output, including denormalization, to obtain the final physical fields; and (iv) it calculates the relative mass residual from the newly predicted velocity field. This loop continues until the calculated residual exceeds the pre-defined threshold, at which point the Predictor halts and returns control to the Orchestrator, reporting the final timestep it successfully reached.

\subsubsection{Solver interface and data exchange}
\label{sec:solver_interface}

The Solver interface acts as the crucial bridge between the Python-based ML environment and the C++-based OpenFOAM solver, encapsulating all direct interactions. It programmatically executes the OpenFOAM solver, both for the initial data generation and for the intermediate corrections during the hybrid loop. Furthermore, it automates the bidirectional data flow required to couple the two environments:

\textbf{CFD-to-ML Conversion (\texttt{foamToNumpy}):} To prepare data for the neural network, the framework leverages the open-source \texttt{Ofpp} library \cite{ofpp}. This utility efficiently parses OpenFOAM's structured text files and converts the high-fidelity field data into standard NumPy arrays \cite{numpy}, making it readily accessible for training and pre-processing within the Python ecosystem.

\textbf{ML-to-CFD Conversion (\texttt{numpyToFoam}):} To transfer ML predictions back to the solver, a custom utility was developed. This tool takes the model's output as NumPy arrays and uses a regular-expression-based method \cite{regex} to surgically insert the numerical data into the correct \texttt{internalField} section of OpenFOAM's dictionary-style files. This process ensures the file syntax is preserved for a seamless restart. This utility also orchestrates the critical steps for ensuring physical consistency, initiating the re-calculation of the density field ($\rho$) and triggering the mass flux ($\phi$) adjustment routine discussed in the following section. The algorithmic logic is detailed in Algorithm \ref{alg:numpyToFoam}.

\begin{algorithm}[htbp]
    \caption{ML-to-CFD data conversion (\texttt{numpyToFoam})}
    \label{alg:numpyToFoam}
    \begin{algorithmic}[1]
        \State \textbf{Input:} Predicted field variables  $\mathbf{X}$, OpenFOAM configuration, directory paths, CFD reference time $t_{\text{CFD}}$, ML prediction time $t_{\text{ML}}$
        \State Identify the existing CFD time directory: $t_{\text{CFD}}$
        \State Create a new time directory $t_{\text{ML}}$ by copying data from $t_{\text{CFD}}$
        \State Update all location entries in header files to reflect $t_{\text{ML}}$
        
        \For{each field variable $x \in\mathbf{X}$}
            \State Load predicted NumPy array: $x_{\text{ML}} = \texttt{np.load}(x_{t_{\text{ML}}})$
            \State Convert NumPy data to string format using \texttt{parse\_numpy}:
            \[
            x_{\text{foam}} = \texttt{parse\_numpy}(x_{\text{ML}})
            \]
            \State Apply regular expression (regex) to replace the existing data with ML predicted one
            \[
            \texttt{re.sub\_foam\_content}(x_{\text{foam}})
            \]
        \EndFor
        
        \State Compute density $\rho$:
        \[
        \rho = \frac{p(t_{\text{CFD}}) \cdot W}{R \cdot T(t_{\text{ML}})}
        \]
        where $W = 0.02896$ kg/mol and $R = 8.314$ J/mol·K
        
        \State Insert $\rho$ field into the corresponding file in $t_{\text{ML}}$ directory
        
        \State Correct flux field $\phi$: \Comment{By calling external utility} 
        \[
        \texttt{adjustPhiML -case solver\_dir -time t\_ML}
        \]
        
        \State \textbf{Output:} Updated OpenFOAM files at time $t_{\text{ML}}$ with ML-predicted and derived fields
    \end{algorithmic}
\end{algorithm}

\subsection{Adaptive control and physics-informed data handling}
\label{sec:adaptive_control}

One of the cores of our method is a multi-stage, adaptive control loop that ensures physical consistency at every stage of the simulation. This process involves three key stages: pre-emptive physics enforcement on the input data, a dynamic switching criterion to monitor predictions, and a post-prediction correction to guarantee mass conservation.

\subsubsection{A priori boundary condition enforcement}
\begin{figure*}[htbp]
    \centering
    \includegraphics[width=\textwidth]{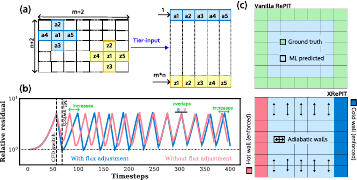}
    \caption{\textbf{Physics-aware components of the timestep-coupled hybrid workflow.} \textbf{(a)} The physics-inspired tiered stencil input structure. The input for each cell is a feature vector containing its own value and the values of its immediate neighbors, providing the neural network with local spatial context analogous to a finite volume discretization.
    \textbf{(b)} Importance of the a posteriori mass flux correction. The plot demonstrates that applying the \texttt{adjustPhiML} utility to enforce mass conservation on the ML-predicted velocity field significantly stabilizes the simulation, enabling longer ML rollouts. \textbf{(c)} A priori boundary condition enforcement. Before being passed to the ML model, the domain is padded with an extra layer of cells that are filled with the appropriate physical boundary conditions, making the surrogate inherently aware of the domain constraints.}
    \label{fig:methods_fig2}
\end{figure*}

To ensure the ML model respects the physical boundaries of the domain, we enforce boundary conditions directly at the data pre-processing stage, before the data enters the neural network. As illustrated in \textbf{Fig.~\ref{fig:methods_fig2}(c)} and in Algorithm \ref{alg:bc_enforcement}, the input fields are padded with an extra layer of cells that are filled with the appropriate Dirichlet (fixed value) or Neumann (zero-gradient) conditions. This embeds the boundary physics directly into the input tensor, making the model inherently aware of the domain constraints.

\begin{algorithm}[htbp]
    \caption{A priori boundary condition enforcement}
    \label{alg:bc_enforcement}
    \begin{algorithmic}[1]
        \State \textbf{Assume:} Top and bottom walls are adiabatic (Neumann), left and right walls are Dirichlet; no-slip if velocity field
        \State \textbf{Input:} 2D field $Z \in \mathbb{R}^{H \times W}$ (e.g., temperature or velocity)
        \State Initialize padded field $Z' \in \mathbb{R}^{(H+2) \times (W+2)}$ with zeros
        
        \For{$i = 0$ to $H-1$}
            \For{$j = 0$ to $W-1$}
                \State $Z'_{i+1,j+1} \gets Z_{i,j}$ \Comment{Copy interior values}
            \EndFor
        \EndFor
        
        \For{$j = 1$ to $W$}
            \State $Z'_{0,j} \gets Z'_{1,j}$ \Comment{Top adiabatic: $\partial Z / \partial y = 0$}
            \State $Z'_{H+1,j} \gets Z'_{H,j}$ \Comment{Bottom adiabatic}
        \EndFor
        
        \For{$i = 0$ to $H+1$}
            \State $Z'_{i,0} \gets Z_{\text{left}}$ \Comment{Left Dirichlet (or zero for no-slip)}
            \State $Z'_{i,W+1} \gets Z_{\text{right}}$ \Comment{Right Dirichlet}
        \EndFor
        
        \State \textbf{Output:} Padded field $Z' \in \mathbb{R}^{(H+2) \times (W+2)}$ with physical boundary conditions
    \end{algorithmic}
\end{algorithm}

    \subsubsection{Residual-guided switching logic}
Once the model is trained and the auto-regressive prediction phase begins, its validity is continuously monitored using a switching criterion based on the physical residuals of the governing equations, a technique that has proven effective as an error indicator in computational physics~\cite{hajibeygi2011residueAsErrorMetric, residueAsAccuracy}. At each ML-predicted step, we compute a mass conservation residual (Eq. ~\eqref{eq:R_mass}). As mentioned in Section ~\ref{subsec:vanilla_repit}, this is normalized to obtain a relative residual as in Eq. ~\eqref{eq:relative_residual_limit}.

It is worth noting that the normalization reference $R_{mass}^{t_0}$ in Eq. ~\eqref{eq:relative_residual_limit} is computed from the final timestep of the initial CFD training window and remains fixed throughout the simulation. Since the present study begins from the transient regime itself, this reference is representative of the flow dynamics at the point where ML inference takes over, and the relative residual therefore provides a meaningful measure of how much the ML predictions deviate from the expected physics at that stage. One might question whether this fixed baseline becomes inappropriate as the flow evolves toward a statistically different regime. However, the self-correcting nature of the hybrid loop inherently mitigates this concern to some extent: even if the reference causes the threshold to be crossed slightly earlier or later during a particular ML rollout, the subsequent CFD correction—comprising a short solver burst and online transfer learning—re-anchors the solution in high-fidelity physics before the next rollout begins. Any transient inaccuracy in the switching decision during one cycle is therefore corrected by the intermediate CFD intervention, preventing errors from accumulating across cycles. We acknowledge, however, that a fixed reference may not be equally effective for all flow configurations, particularly those involving sharp regime transitions where the baseline statistics become unrepresentative. To address such scenarios, the framework architecture supports two straightforward extensions: (i) replacing the fixed reference with a multi-timestep averaged scaler computed over a specified window of recent timesteps, or (ii) dynamically updating the reference using the residual computed from the fresh CFD solution generated at each switching event. Both strategies require only a minor modification to the normalization routine and are readily accessible within the existing codebase. Since the fixed-baseline approach proved sufficient for stable operation over 10,000 timesteps across cases studied in this work (see Section ~\ref{sec:results}), we defer the exploration of adaptive reference strategies to future case studies where such refinement becomes necessary.

\subsubsection{A posteriori mass flux correction}
Upon crossing the residual threshold, the framework initiates a sequence to reground the simulation in physics. Crucially, before the CFD solver is reinvoked, the predicted velocity field is passed to a custom C++ utility, \texttt{adjustPhiML}. This tool recalculates the mass flux ($\phi$) and projects it onto a divergence-free space, enforcing mass conservation (see Algorithm \ref{alg:PhiAdjustment}). This step is not a minor correction; as shown in \textbf{Fig.~\ref{fig:methods_fig2}(b)}, it ensures that the velocity field is physically consistent before the solver begins, enabling much longer ML rollouts in subsequent cycles. Following this consistency enforcement, the OpenFOAM solver is run for a short burst (e.g., 10 timesteps), and the final three timesteps are used for targeted online transfer learning, which leverages the data-efficient nature of the tiered-input (\textbf{Fig.~\ref{fig:methods_fig2}(a)}) FVMN architecture~\cite{jeon2022FVMN}.

\begin{algorithm}[htbp]
    \caption{A posteriori mass flux correction (\texttt{adjustPhiML})}
    \label{alg:PhiAdjustment}
    \begin{algorithmic}[1]
        \State \textbf{Input:} ML predicted velocity field \(\mathbf{u}\), pressure field \(p\) from latest CFD timestep, calculated density field \(\rho\), mesh and latest ML timestep
        \State Parse command-line arguments: \texttt{-time <T>} or \texttt{-latestTime}
        \State Initialize OpenFOAM case environment: \texttt{setRootCase}, \texttt{createTime}, \texttt{createMesh}
        \State Select the latest ML predicted time directory $(t_{\text{ML}})$ from user input
        \State Load fields from disk:
        \[
        \mathbf{u}(\mathbf{x}, t_{\text{ML}}), \quad p(\mathbf{x}, t), \quad \rho(\mathbf{x}, t_{\text{ML}})
        \]
        \State Compute initial flux using density-weighted velocity:
        \[
        \phi = \texttt{fvc::flux}(\rho \mathbf{u}) = \int_S \rho \mathbf{u} \cdot \mathbf{n} \, dS
        \]
        \State Call correction routine:\Comment{Adjust \(\phi\) to ensure mass conservation}
        \[
        \phi \leftarrow \texttt{adjustPhi}(\phi, \mathbf{u}, p)
        \]
        
        \State Write corrected \(\phi\) field to disk
        \State \textbf{Output:} Updated surface flux field \(\phi\) for ML predicted time directory
    \end{algorithmic}
\end{algorithm}

\subsection{Model implementations and training}
\label{sec:nn_method}
Within the finite-volume-based framework described in Section ~\ref{subsec:fvmn}, we benchmarked two specific model implementations that differ in the type of sub-network used.

\begin{itemize}
    \item \textbf{FVMN:} Employs a standard multi-layer perceptron(MLP) as the core processing block for each sub-network. The improvements in the network workflow and architecture, along with the hyper-parameters, can be found in Appendix \ref{sup_subsec:fvmn}. 
    \item \textbf{FVFNO:} Employs a Fourier neural operator as the core processing block in place of a normal MLP, compared to FVMN, which is designed to capture a wider range of spatial dependencies.
    In Appendix \ref{sup_subsec:fvfno}, the mathematical formulation and explanations of each step of the calculation are outlined along with the hyper-parameters' values.
\end{itemize}

Both the initial training and the subsequent online transfer learning cycles were performed using a dataset of only \textbf{three} consecutive high-fidelity timesteps generated by the CFD solver. Despite this minimal training data, the hybrid method achieved long and stable prediction rollouts as suggested in \textbf{Fig. \ref{fig:results_fig3}}.

\subsection{Performance and error metrics}
\label{sec:metrics}

To ensure clarity and reproducibility, the performance and accuracy of the hybrid simulations were quantified using a consistent set of formally defined metrics.

\subsubsection{Performance metrics}
The computational performance was evaluated using a speedup factor, $\psi$, defined as the ratio of the wall-clock time required for a pure CFD simulation to the total time for the hybrid simulation:
\begin{equation}
    \psi \approx \frac{T_{CFD}}{T_{Hybrid}}
\end{equation}
Crucially, $T_{Hybrid}$ is the total wall-clock time and includes all computational overheads associated with the hybrid method: ML inference, CFD runs, and transfer learning updates.

The precise calculation for the speedup factor, accounting for each component of the hybrid loop, is given by:
\begin{equation}
    \label{eq:acceleration}
    \psi = \frac{N \cdot t_{CFD}}{n_{CFD} \cdot t_{CFD} + n_{ML} \cdot t_{ML} + n_{switch} \cdot t_{up}}
\end{equation}
where $N$ is the total number of timesteps in the simulation, $t_{CFD}$ is the average time per timestep for the OpenFOAM solver, $t_{ML}$ is the average time for a single ML inference step, $t_{up}$ is the average time for one online transfer learning update, and $n_{CFD}$, $n_{ML}$, and $n_{switch}$ are the total counts of CFD steps, ML steps, and ML-to-CFD switches in the hybrid simulation, respectively.  

Additionally, it is confirmed in this study that multi-core CPU parallelization offers no benefit over a single-core calculation for this less extensive natural convection flow problem in AMD EPYC 9554 256 core engine. Hence, all timing results reported here are based on single-core execution for CFD and GPU parallelization in NVIDIA A100(40GB) for ML training and inference. Also, extended software and hardware information can be found in Appendix ~\ref{sup_sec:hardsoft_info}

\subsubsection{Accuracy metrics}
To rigorously evaluate the performance of the hybrid simulation, we quantified the accuracy by comparing the predicted field variables, $\hat{Z}$, against the ground truth high-fidelity CFD data, $Z$. We employed a suite of four distinct error metrics, each selected to probe a different facet of the model's predictive fidelity, from global accuracy to worst-case local deviations.

\begin{itemize}
    \item \textbf{Relative L\textsubscript{2} Error:} This metric provides a holistic measure of field-level accuracy by quantifying the normalized Euclidean distance between the predicted and true fields. It offers a concise, global assessment of the model's performance.
    \begin{equation}
        \text{Relative L\textsubscript{2} Error} = \frac{\left\|Z - \hat{Z}\right\|_2}{\left\|Z\right\|_2}
    \end{equation}

    \item \textbf{MSE:} As the average of squared differences, the MSE is particularly sensitive to large deviations. A consistently low MSE serves as a strong indicator of model robustness, confirming the absence of significant, large-scale prediction failures.
    \begin{equation}
        \text{MSE} = \frac{1}{N_{\text{cells}}} \sum_{i=1}^{N_{\text{cells}}} (Z_i - \hat{Z}_i)^2
    \end{equation}

    \item \textbf{Mean Absolute Error (MAE):} This metric offers a direct and interpretable measure of the average prediction error across the domain. The temporal stability of the MAE is a critical diagnostic for autoregressive models, as it demonstrates that incremental inaccuracies do not accumulate over long-term rollouts.
    \begin{equation}
        \text{MAE} = \frac{1}{N_{\text{cells}}} \sum_{i=1}^{N_{\text{cells}}} |Z_i - \hat{Z}_i|
    \end{equation}
    
    \item \textbf{Maximum Absolute Error (MaxAE):} Representing the most stringent test of model reliability, the MaxAE identifies the worst-case, pointwise error at any location within the domain. A bounded MaxAE is crucial, as it validates the framework's ability to control and suppress localized error hotspots that could otherwise grow and destabilize the simulation.
    \begin{equation}
        \text{MaxAE} = \max_{i} |Z_i - \hat{Z}_i|
    \end{equation}
\end{itemize}

\section{Results}\label{results}
\label{sec:results}
\subsection{Hybrid framework to overcome catastrophic failure in auto-regressive surrogates}\label{single_training}

\begin{figure*}[htbp]
    \centering
    \includegraphics[width=\textwidth]{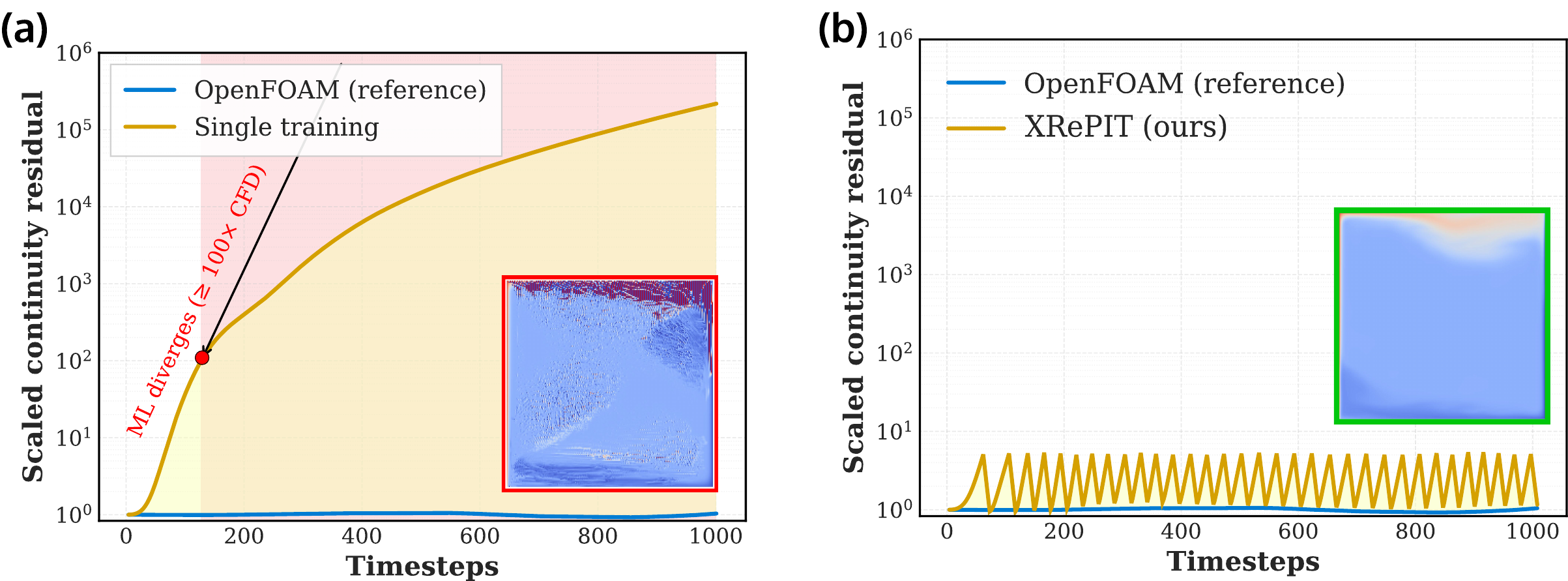}
    \caption{\textbf{Residual comparison between single training and XRePIT workflow.}\textbf{(a)} The failure of a conventional data-driven surrogate. The relative residual of a FVMN-based model grows exponentially during an autoregressive rollout, leading to a distorted, non-physical temperature field at 1,000 timesteps as shown in inset. \textbf{(b)} The stability claim of the proposed method which is brought by the constant monitoring of the residual mass as the flow evolves. The strategy for using residual threshold to quantify the error accumulation made the temperature field stabilize even after 1000 timesteps.}
    \label{fig:results_fig1}
\end{figure*}

As we have already stated, although high-fidelity computational fluid dynamics (CFD) simulations are accurate, they are computationally prohibitive. For example, even to calculate the natural convection with fewer degrees of freedom, it took almost 4000 s just to simulate 100 s of the flow with very relaxed solver settings. This motivates the use of data-driven surrogates. But a critical limitation of purely auto-regressive models is their rapid accumulation of errors. We first demonstrate this failure mode using a standard finite volume method network (FVMN)~\cite{jeon2022FVMN}. As shown in \textbf{Fig.~\ref{fig:results_fig1}(a)}, the relative residual diverges catastrophically, increasing by over five orders of magnitude within just 1,000 timesteps. This numerical instability leads to a complete breakdown of the physical solution, as evidenced by the distorted and non-physical temperature field.

Here, we apply the residual-guided correction strategy, implemented via the XRePIT framework, which synergistically couples neural networks for acceleration with traditional numerical solvers for stabilization. Instead of failing, the hybrid loop effectively contains the error growth. The framework monitors the residual and, when the threshold is breached, invokes the CFD solver to stabilize the prediction before resuming accelerated ML inference. This intervention ensures that the residual never exceeds the defined threshold, allowing the simulation to remain physically grounded and stable over long-term rollouts \textbf{(Fig.~\ref{fig:results_fig1}(b))}. This stability ensures the accurate capture of complex flow structures, a feat unattainable with the standalone auto-regressive model.

Moreover, the choice of mass conservation as the sole switching criterion is grounded in the physics of the studied regime. In buoyancy-driven flows at moderate Rayleigh numbers, the errors in the predicted velocity field first manifest as non-zero divergence, which then propagates into momentum and energy through incorrect pressure gradients and erroneous advective fluxes respectively. The mass residual therefore serves as an upstream indicator of error accumulation. We verify this empirically in two ways. First, Fig. 6(a) confirms that in standalone surrogate rollouts, the mass residual diverges earliest and most steeply, consistently preceding the growth in momentum and energy residuals. Second, Table ~\ref{tab:overall_performance_case1} establishes a monotonic correlation between the mass residual threshold and all field error metrics: tightening the threshold from 100 to 5 reduces L2(T) from 1.83e-3 to 8.29e-4 and L2(U) from 0.558 to 0.296, demonstrating that bounding the mass residual implicitly bounds the momentum and energy errors for this flow regime.
We acknowledge that this empirical coupling is regime-specific. In flows where momentum or energy source terms dominate (e.g., strongly advective, reactive, or turbulent regimes), these fields may diverge while continuity remains approximately satisfied. For such cases, the framework is designed to accommodate multi-residual switching.

\begin{figure*}[htbp]
    \centering
    \includegraphics[width=\textwidth]{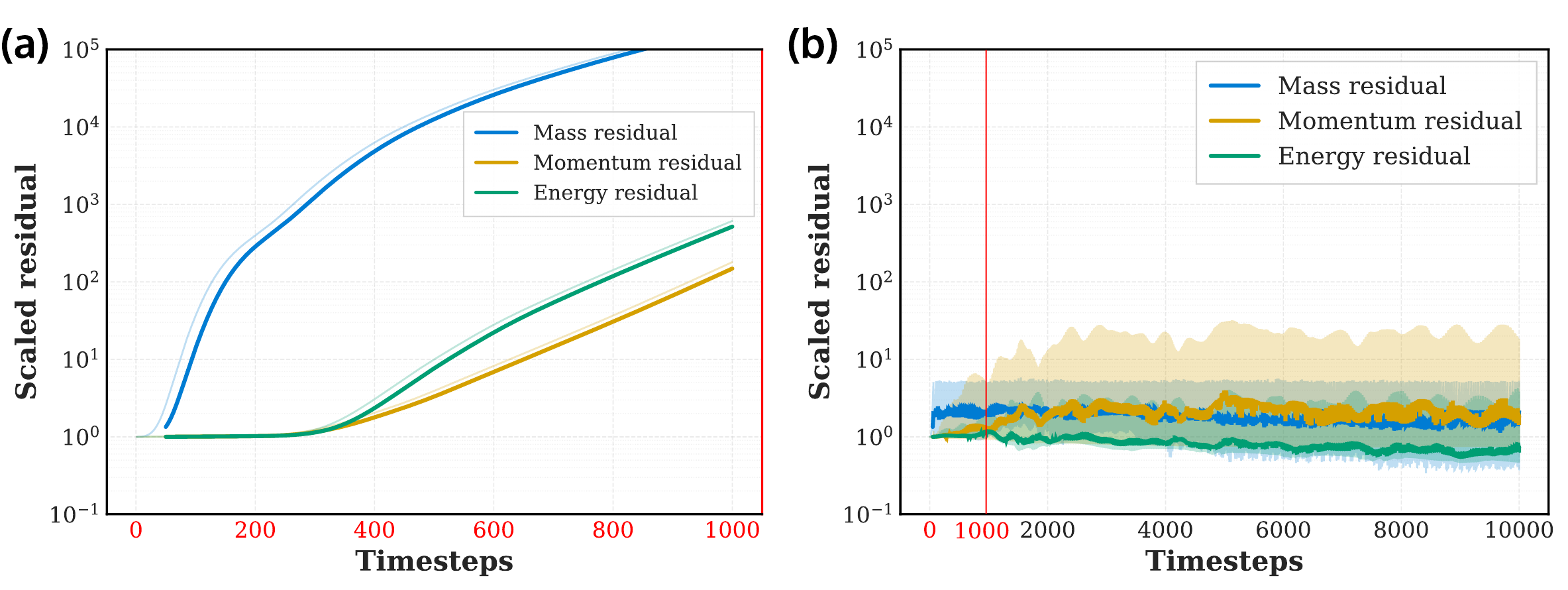}
    \caption{\textbf{Long term stability of mass, momentum, and energy residuals within XRePIT.}\textbf{(a)} In the standalone surrogate, all three residuals 
    diverge; the mass residual provides the earliest and steepest divergence signal, 
    consistently preceding the growth in momentum and energy. \textbf{(b)} Monitoring 
    only the mass residual within XRePIT is sufficient to keep all three residuals 
    bounded over 10,000 timesteps, demonstrating that continuity-based switching 
    implicitly stabilizes the coupled momentum and energy fields.}
    \label{fig:results_fig2_all_residuals}
\end{figure*}

\subsection{Tunable performance of the hybrid methodology}
\label{subsec:hyperparameters}

After realizing error stabilization for 1000 time steps, we used the XRePIT framework to test beyond this region, and we will show in the subsequent sections how sustained the results become even after thousands of time steps. But before that, we need to explore the capacity of the timestep-coupled hybrid method for tunable acceleration and accuracy. The trade-off between simulation speed and physical fidelity is governed by two key hyperparameters within the hybrid logic: the relative residual threshold, which dictates the frequency of CFD corrections, and the number of transfer-learning epochs, which controls the extent of model adaptation. A systematic analysis of these parameters reveals a clear and controllable performance landscape, as detailed in \textbf{Table~\ref{tab:overall_performance_case1}}.

\begin{figure*}[htbp]
\centering
\includegraphics[width=\textwidth]{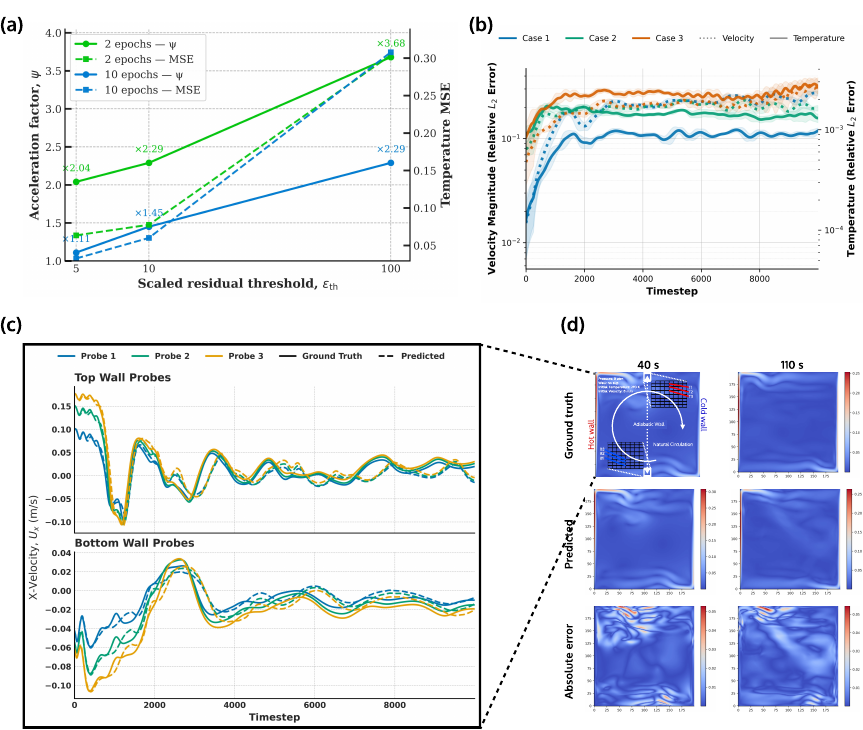}
\caption{\textbf{XRePIT performance and long-term stability analysis.} \textbf{(a)} The relationship between the acceleration factor, MSE for temperature, the relative residual threshold, and the number of transfer learning epochs. It illustrates how the hyperparameters can be tuned to balance computational speed against predictive accuracy. \textbf{(b)} The relative L\textsubscript{2} error over time for temperature (right) and velocity magnitude (left) under different physical boundary conditions. This demonstrates the framework's robustness, as the prediction errors for both quantities are consistently maintained at a minimal level (less than 1\% for temperature and around 10\% for low magnitude velocity). \textbf{(c)} Line plots comparing the values predicted by XRePIT against the ground truth at the probe locations defined in (d). The results show that the hybrid simulation accurately follows the same physical trends as the ground truth solver, even during very long-term simulations, confirming the stability. \textbf{(d)} A visual comparison between the velocity magnitude fields predicted by XRePIT and the ground truth solver at t = 40s and t = 110s. The corresponding absolute error fields remain low (in the range of 1e-2), demonstrating high fidelity even after 10,000 timesteps in the hybrid regime.}
\label{fig:results_fig2}
\end{figure*}

The results show a relationship between the residual threshold, acceleration, and error. Increasing the residual threshold allows for longer, uninterrupted ML rollouts. This reduces the frequency of costly CFD corrections, directly boosting the acceleration factor ($\psi$) at the expense of a quantifiable increase in the Mean Squared Error (MSE) (see \textbf{Fig.~\ref{fig:results_fig2}(a)}). For instance, in the 2-epoch configuration, relaxing the threshold from 5 to 100 reduced the number of solver interventions ($n_{switch}$) by more than half, from 343 to 170. This reduction in computational overhead elevates the acceleration from \textbf{2.04x} to a remarkable \textbf{3.68x}, while the mean squared error (MSE) for the temperature field increased by nearly fivefold, from 0.063 to 0.305 (\textbf{Table~\ref{tab:overall_performance_case1}(b)}).

Our hyperparameter study revealed a critical, non-linear interaction between transfer-learning epochs and the residual threshold \textbf{(Fig.~\ref{fig:results_fig2}(a))}. We found that at higher residual thresholds, the 10-epoch configuration offers no significant accuracy benefit and can even perform worse than the 2-epoch run. This is because the model attempts to fit longer to the more erroneous dynamics allowed by the high threshold. This lack of benefit is compounded by a steep computational cost: the 10-epoch update time ($t_{up}$) is $\sim$5x longer than the 2-epoch one. This analysis makes the trade-off clear: the 2-epoch model provides comparable fidelity with far greater acceleration. We thus identified the 2-epoch, 5-residual configuration as the clear optimal balance, and we use this ``2-5 variant'' for all subsequent studies. This in-depth analysis was repeated for the other boundary cases, yielding the same conclusion (Tables \ref{sup_tab:overall_performance_case2} \& \ref{sup_tab:overall_performance_case3}).

While this configuration proves optimal for the tested cases, scaling to higher Reynolds numbers or more chaotic flows would likely necessitate adjustments. In scenarios with stronger transients or sharper gradients, we anticipate that a tighter residual threshold would be required to capture rapidly diverging physics earlier. Conversely, for highly chaotic flows where the manifold is more complex, increasing the number of transfer-learning epochs might become necessary to allow the surrogate sufficient capacity to adapt to the richer spectrum of dynamics, even at the cost of some acceleration.

\begin{sidewaystable*}[htbp]
  \centering
  \caption{Results obtained after treating hybridization adaptive parameters as hyperparameters.}
  \label{tab:overall_performance_case1}

  \begin{threeparttable}
  \footnotesize

  \begin{tablenotes}\footnotesize
    \item \textbf{Res.}: Relative residual threshold.
    \textbf{$t_{\mathrm{CFD}}$}: Solver time per timestep.
    \textbf{$t_{\mathrm{ML}}$}: ML inference time per timestep.
    \textbf{$t_{\mathrm{up}}$}: Parameter update time per switch.
    \textbf{$n_{\mathrm{switch}}$}: Number of ML--CFD switches.
    \textbf{$n_{\mathrm{CFD}}$}: Total number of CFD timesteps.
    \textbf{$n_{\mathrm{ML}}$}: Total number of ML timesteps.
    \textbf{OpenFOAM (s)}: Total wall-clock time for the CFD-only simulation.
    \textbf{XRePIT (s)}: Total wall-clock time for the hybrid simulation.
    \textbf{$\psi$}: Speedup factor.
    \textbf{$t_{\mathrm{avg.\,switch}}$}: Average timesteps per switch.
  \end{tablenotes}

  \vspace{0.75ex}
  \textbf{(a) Acceleration analysis on epoch and residual-threshold configurations.}
  \vspace{0.5ex}

  \setlength{\tabcolsep}{3.5pt}
  \renewcommand{\arraystretch}{1.1}
  \begin{tabular}{@{}cccccccccccc@{}}
    \toprule
    \textbf{Epochs} & \textbf{Res.} &
    $\boldsymbol{t_{\mathrm{CFD}}}$ &
    $\boldsymbol{t_{\mathrm{ML}}}$ &
    $\boldsymbol{t_{\mathrm{up}}}$ &
    $\boldsymbol{n_{\mathrm{switch}}}$ &
    $\boldsymbol{n_{\mathrm{CFD}}}$ &
    $\boldsymbol{n_{\mathrm{ML}}}$ &
    \textbf{OpenFOAM (s)} &
    \textbf{XRePIT (s)} &
    $\boldsymbol{\psi}$ &
    $\boldsymbol{t_{\mathrm{avg.\,switch}}}$ \\
    \midrule
    2  &   5 & 0.42 & 0.026 & 1.30 & 343 & 3423 & 6585 & 4252.8 & 2075.8 & 2.04 & 19.19 \\
    2  &  10 & 0.43 & 0.027 & 1.26 & 293 & 2923 & 7077 & 4319.6 & 1828.0 & 2.36 & 24.15 \\
    2  & 100 & 0.43 & 0.026 & 1.31 & 170 & 1693 & 8307 & 4380.4 & 1188.8 & 3.68 & 48.86 \\
    10 &   5 & 0.42 & 0.025 & 6.66 & 305 & 3043 & 6957 & 4247.5 & 3504.7 & 1.21 & 22.80 \\
    10 &  10 & 0.42 & 0.025 & 6.21 & 260 & 2593 & 7407 & 4228.2 & 2901.9 & 1.45 & 28.48 \\
    10 & 100 & 0.43 & 0.026 & 6.22 & 160 & 1593 & 8407 & 4392.4 & 1915.2 & 2.29 & 52.54 \\
    \bottomrule
  \end{tabular}

  \vspace{1.25ex}
  \textbf{(b) Time-averaged spatial error metrics (Case 1).}
  \vspace{0.5ex}

  \setlength{\tabcolsep}{4pt}
  \renewcommand{\arraystretch}{1.1}
  \begin{tabular}{@{}cccccccccc@{}}
    \toprule
    \textbf{Epochs} & \textbf{Res.} &
    \textbf{L$_2$(T)} & \textbf{MSE(T)} & \textbf{MAE(T)} & \textbf{MaxAE(T)} &
    \textbf{L$_2$(U)} & \textbf{MSE(U)} & \textbf{MAE(U)} & \textbf{MaxAE(U)} \\
    \midrule
    2  &   5 & 8.29e$-$4 & 0.063 & 0.151 & 1.60 & 0.296 & 1.27e$-$4 & 0.008 & 0.067 \\
    2  &  10 & 9.27e$-$4 & 0.078 & 0.166 & 1.78 & 0.335 & 1.63e$-$4 & 0.009 & 0.071 \\
    2  & 100 & 1.83e$-$3 & 0.305 & 0.345 & 2.86 & 0.558 & 4.56e$-$4 & 0.016 & 0.104 \\
    10 &   5 & 5.96e$-$4 & 0.033 & 0.107 & 1.35 & 0.213 & 6.55e$-$5 & 0.006 & 0.051 \\
    10 &  10 & 8.03e$-$4 & 0.060 & 0.137 & 1.74 & 0.283 & 1.15e$-$4 & 0.008 & 0.062 \\
    10 & 100 & 1.75e$-$3 & 0.306 & 0.298 & 3.00 & 0.591 & 5.12e$-$4 & 0.016 & 0.105 \\
    \bottomrule
  \end{tabular}

  \end{threeparttable}
\end{sidewaystable*}

\subsection{Boundary condition adaptation of the hybrid method}
\label{subsec:generalization}

A critical test for any hybrid simulation strategy is whether it can adapt when the physical setup changes, without incurring the cost of retraining a surrogate from scratch for every new scenario. In this work, we use ``generalization'' in a sense: adaptation across boundary conditions via online transfer learning under a fixed geometry and numerical configuration. We evaluate this by deploying the hybrid loop on two additional cases (Case 2 and Case 3; Section~\ref{subsubsec:domain_and_case}) whose thermal boundary conditions differ from the baseline (Case 1) used for offline surrogate training.

The same pre-trained neural network from Case 1 initializes the hybrid simulations for Case 2 and Case 3, and the model adapts to the new boundary conditions \textit{only} through the framework's intrinsic online transfer-learning cycles (i.e., no separate offline pretraining is performed for Case 2/3). This directly tests the adaptive capacity of the residual-guided hybrid logic and highlights a practical advantage: rapid deployment across boundary-condition variants with substantially reduced time-to-solution. This systematic evaluation is enabled by our automated workflow, which manages data handling, switching decisions, and solver coupling without manual intervention.

The results show that physics-based correction through CFD fallback stabilizes the rollout while the surrogate adapts online. As quantified in Fig.~\ref{fig:results_fig2}(b), the relative $L_2$ errors for temperature and velocity magnitude remain low across all three cases, despite a characteristic error growth during the initial phase. This early transient reflects the method's ``adaptation lag'' in response to the highly dynamic startup physics, where the rapid evolution of thermal structures initially challenges the surrogate's extrapolation capabilities. However, the residual-based switching mechanism successfully manages this complexity by enforcing frequent CFD corrections, effectively using this period as an online ``warm-up'' to realign the model weights. Consequently, as shown in Appendix Fig.~\ref{sup_fig:error_analysis_cases}, the error metrics stabilize rapidly and stay bounded over the remainder of the 10{,}000-timestep rollout.

\begin{figure*}[htbp]
    \centering
    \includegraphics[width=\textwidth]{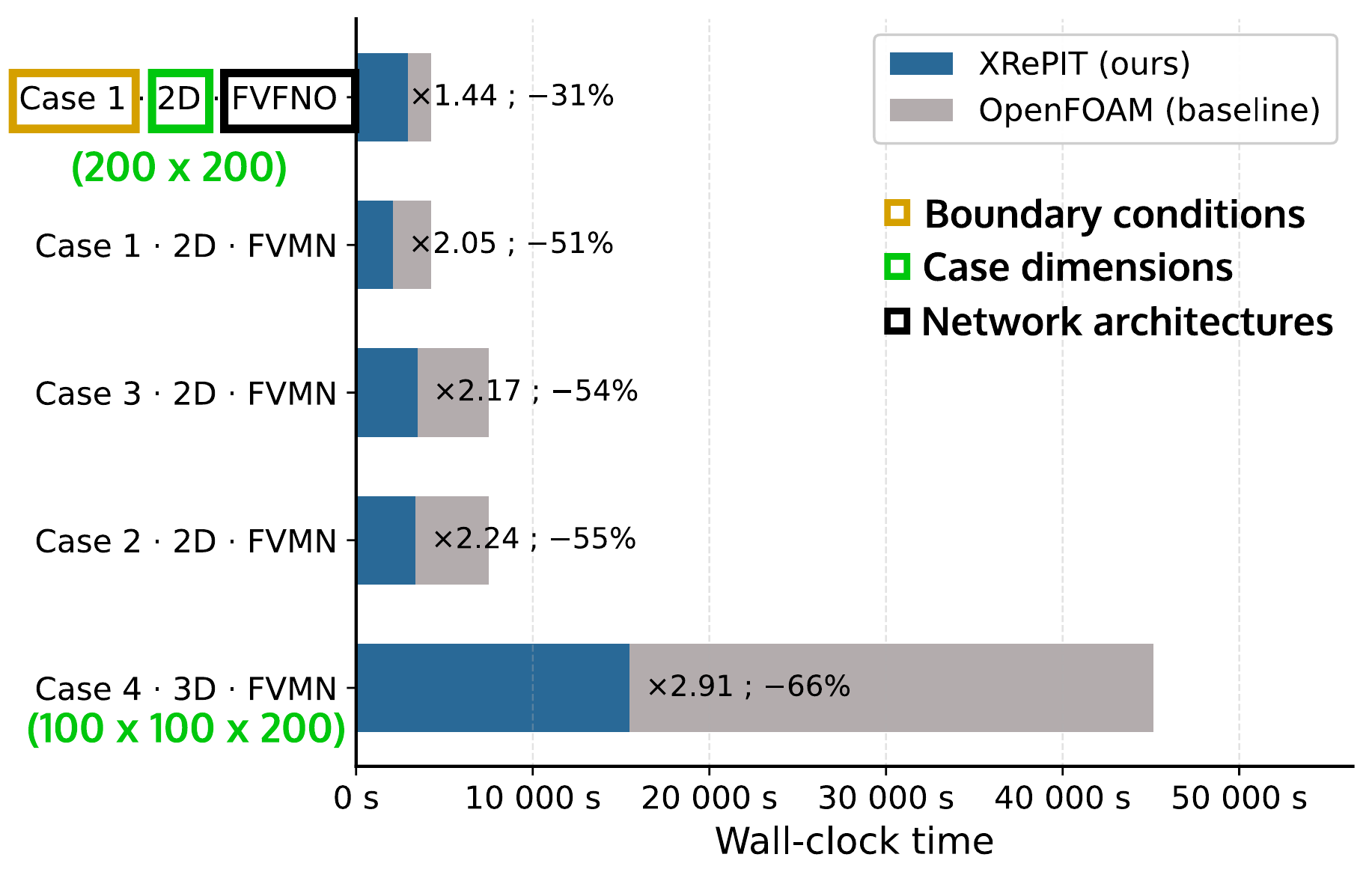}
    \caption{A summary of the wall-clock time comparison between the proposed framework (blue) and the traditional CFD solver (gray) for all cases investigated throughout the study.}
    \label{fig:acceleration_barplot}
\end{figure*}

To probe localized dynamics, we sample six highly active boundary regions and compare the predicted $u_x$ against the CFD reference in Fig.~\ref{fig:results_fig2}(c). Appendix Fig.~\ref{sup_fig:probe_analysis_all} provides analogous comparisons for velocity magnitude and temperature across cases. Qualitatively, the flow fields are also preserved, as illustrated by the velocity snapshots in Fig.~\ref{fig:results_fig2}(d) and the side-by-side comparisons at post-initial (20s), mid (60s), and final (110s) times in Fig.~\ref{sup_fig:visual_comparison}. Overall, these results support the notion that the residual-guided hybrid method provides robust boundary-condition adaptation within the studied buoyancy-driven regime. Importantly, this boundary-condition adaptation is achieved alongside consistent acceleration. As shown in Fig.~\ref{fig:acceleration_barplot}, Case 2 and Case 3 attain speedups of 2.24$\times$ and 2.17$\times$, respectively. We attribute the slightly higher speedups relative to Case 1 to a modest increase in baseline CFD runtime under the new boundary conditions, while ML inference cost remains essentially unchanged. We emphasize that this ``generalization'' claim is limited to online adaptation across boundary conditions for the same configuration.

We can see in Tables~\ref{tab:overall_performance_case1}(b),~\ref{sup_tab:overall_performance_case2},~\ref{sup_tab:overall_performance_case3}, the relative $L_2$ errors for velocity magnitude appear large (up to ${\sim}30\%$). We highlight this headline can be misleading for buoyancy-driven flows. The relative $L_2$ norm divides by $\|\mathbf{u}\|_2$ over the entire field at each timestep, which is dominated by large quiescent regions in the cavity interior where $|\mathbf{u}|$ is very low; consequently, even small absolute deviations are amplified into large relative values. In the same tables, we can see absolute metrics tell a different story: MAE remains below 0.016\,m/s and MSE stays at $\mathcal{O}(10^{-4})$\,(m/s)$^2$. Measured against the boundary-layer velocities of $\mathcal{O}(10^{-1})$\,m/s, the pointwise probe comparisons at six dynamically active boundary-layer locations (Fig.~\ref{sup_fig:probe_analysis_all}) show the hybrid predictions closely tracking the CFD reference over 10,000 timesteps, confirming that the method is accurate where the physics lives; deviations grow slightly only in the low-velocity interior where momentum transport is negligible.

 For reference, these temperature deviations (${\sim}0.5$--$1$\,K) are comparable to standard thermocouple measurement uncertainty (${\pm}1$--$2$\,K) and well within the sensitivity range of RANS turbulence-model selection, both of which are routinely accepted in engineering thermal-hydraulic analysis. For the target applications motivating this work (thermal-hydraulic monitoring in SMRs, digital-twin-based design optimization, and data-driven control) the primary quantities of interest are temperature evolution and macroscopic flow topology over long horizons. Temperature relative $L_2$ errors remain below $\mathcal{O}(10^{-3})$ across all cases, and qualitative flow structure is preserved throughout (Figs.~\ref{fig:results_fig2},~\ref{fig:results_fig4},~\ref{fig:iso_surfaces_3D}). For these engineering quantities, the demonstrated accuracy represents an acceptable trade-off for 2--3$\times$ acceleration.

That said, for applications requiring high pointwise velocity fidelity, the 
framework provides several direct paths to tighten accuracy: lowering the 
residual threshold to trigger more frequent corrections, incorporating the 
momentum residual into the switching criterion, or augmenting the surrogate 
loss with physics-informed velocity penalties. These are parameter-level and 
loss-level adjustments within the existing architecture, not structural 
redesigns.


\subsection{ Comparative architecture benchmarking of SciML models}
\label{subsec:neural_architectures}

\begin{figure*}[htbp]
\centering
\includegraphics[width=\textwidth]{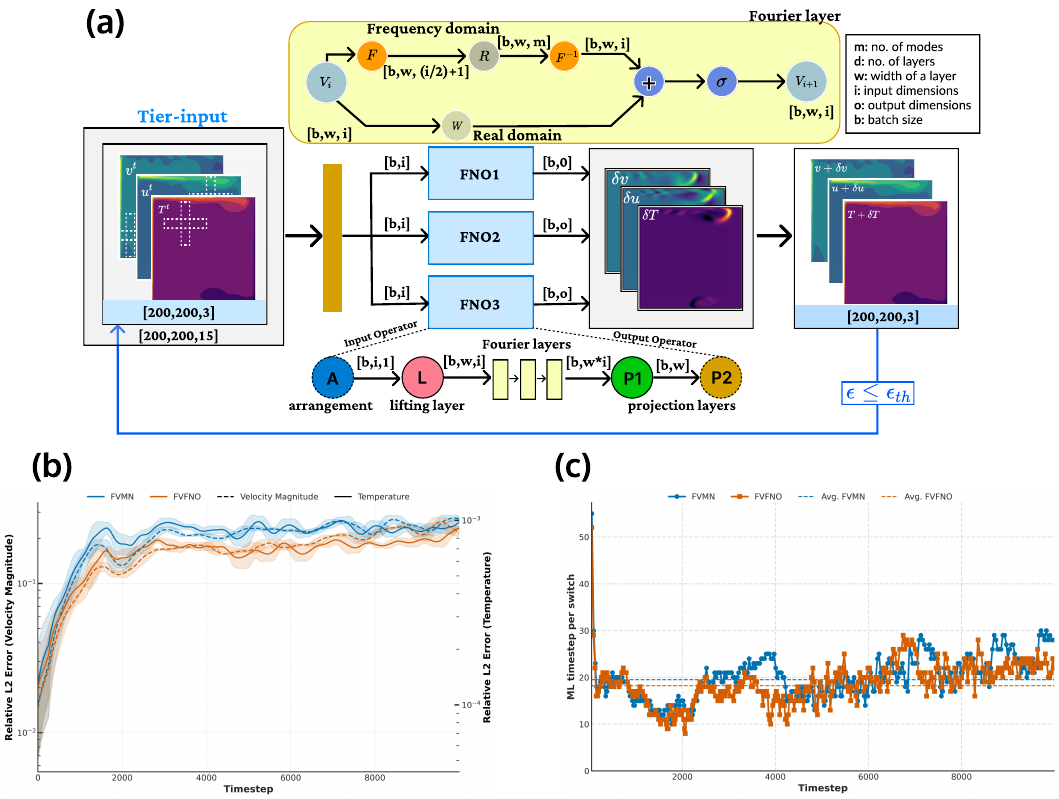}
\caption{\textbf{Architectural extensibility and performance benchmarking in XRePIT.} \textbf{(a)} The architecture features a tier-input system, distinct processing pathways for different physical variables, and a combined derivative loss function to ensure physical consistency. This modular design allows for the FNO block to be replaced with fully connected layers to constitute the baseline FVMN model. \textbf{(b)} Relative L\textsubscript{2} error for velocity magnitude (dotted lines) and temperature (solid lines) over 10,000 timesteps for both the FVMN (blue) and FVFNO (orange) models. Both architectures maintain low, stable error profiles, demonstrating the framework's ability to support different neural network designs without sacrificing physical fidelity. \textbf{(c)} The number of consecutive timesteps predicted by each neural network per switch. The plot shows that for both models, the framework intelligently adapts to flow complexity, taking fewer ML steps during transient phases and increasing the prediction horizon as the flow stabilizes. This highlights the robustness of the residual-guided switching mechanism.}\label{fig:results_fig3}
\end{figure*}

Having established the framework's stability, we next investigate its architectural flexibility. Is the hybrid stability we observed unique to the FVMN, or is the XRePIT workflow truly a ``plug-and-play" tool? To answer this, we introduce a conceptually different, more complex surrogate: a novel \textbf{FVFNO} (\textbf{Fig.~\ref{fig:results_fig3}(a)}). We then use the XRePIT pipeline to benchmark it head-to-head against the original, multi-layer perceptron FVMN.

The results immediately confirm that the framework's stability is not architecture-dependent. Both the FVMN and the FVFNO maintain low, stable error profiles over the entire 10,000-timestep run, with relative L\textsubscript{2} errors in the $10^{-3}$ range for temperature (\textbf{Fig.~\ref{fig:results_fig3}(b)}).

Another significant finding appears in the adaptive switching behavior. \textbf{Fig.~\ref{fig:results_fig3}(c)} shows that the framework's control logic--taking fewer ML steps during rapid transients and more as the flow stabilizes—is nearly identical for both models. This proves that if the flow is not changing rapidly, the residuals also don't increase dramatically; as a result, we would have increased ML prediction timesteps per switch.

This benchmarking capability, however, reveals a critical performance trade-off. While the FVFNO is more accurate, its architectural complexity (requiring fast Fourier transforms and its counterpart) results in a $\sim$4.3x higher inference time (0.112s vs 0.026s). This computational overhead translates directly to a meager \textbf{1.44x} speedup, which is far less than the FVMN's \textbf{2.04x} speedup (\textbf{Fig.~\ref{fig:acceleration_barplot}}).

Our analysis thus demonstrates a key principle for practical hybrid simulation: a more complex, slightly more accurate model (FVFNO) can be an objectively worse choice when acceleration is the goal. The XRePIT framework provides the essential tool for this ``apples-to-apples" comparison, making such systematic optimization of accuracy versus speed attainable.

\subsection{Scalability of the hybrid method to three-dimensional flows}
\label{subsec:scalability}

The ultimate test for any CFD acceleration strategy is its performance on three-dimensional simulations, where computational costs become prohibitively high and the complexity of flow physics increases dramatically. This domain is where acceleration is most needed, and it is here that we provide the final and most critical validation of the timestep-coupled hybrid method.

\begin{figure*}[htbp]
\centering
\includegraphics[width=\textwidth]{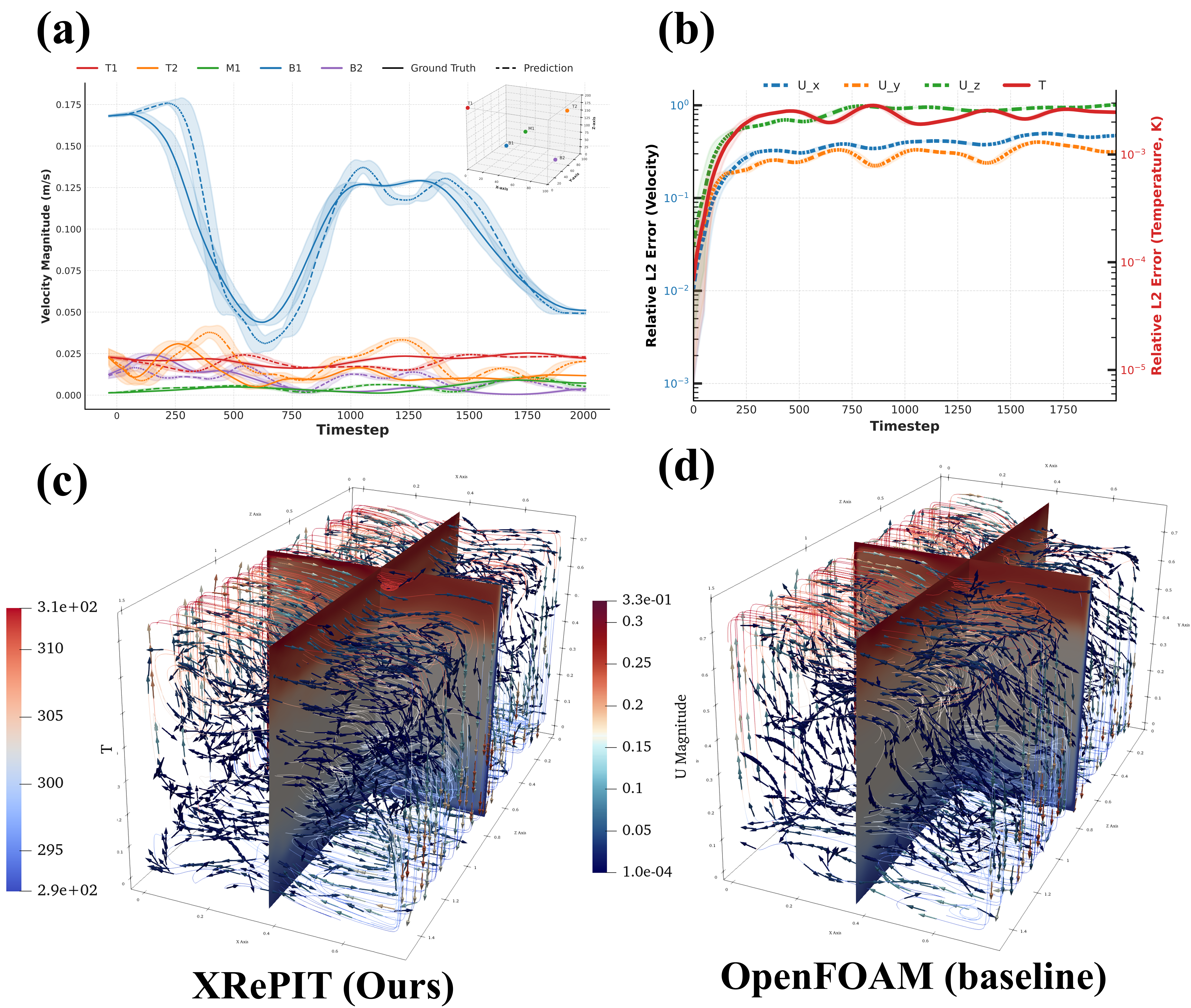}
\caption{\textbf{Scalability and performance of the XRePIT framework in a 3D simulation.}
\textbf{(a)} Comparison of the temporal evolution of a flow variable at a representative probe point between the ground truth (blue) and the XRePIT hybrid prediction (orange), demonstrating long-term trend agreement. \textbf{(b)} Domain-wide MSE for the three velocity components (left) and temperature (right) over 2,000 timesteps, showing error stabilization after an initial transient period. \textbf{(c,d)} Qualitative comparison of the 3D flow field at 30s. The visualization shows streamlines colored by velocity magnitude for the ground truth OpenFOAM simulation  and the XRePIT hybrid simulation, confirming that the complex flow structures are accurately reproduced.}\label{fig:results_fig4}
\end{figure*}

To rigorously assess the method's performance in this scenario, we first conducted a quantitative analysis at semi-randomly selected probe locations within the domain (as inset in \textbf{Fig.~\ref{fig:results_fig4}(a)}). From which we observed how the flow variables at these locations evolve over time. Based on our comparison with the ground truth values, we confirmed that constantly checking on the residual value solves the problem of error accumulation in higher dimensions too. This conclusion is reached from the probes comparison for velocity magnitude in (\textbf{Fig.~\ref{fig:results_fig4}(a)}) and its components along with temperature field in the Fig. \ref{sup_fig:probe_analysis_3d}. 

For a comparable analysis we analyzed the relative L\textsubscript{2} for this case too and found out the results are quite similar to that of the 2D cases (\textbf{Fig.~\ref{fig:results_fig4}(b)}). Also similar to other cases, a steep increase in error for the initial timesteps is seen and in  Fig. \ref{sup_fig:error_analysis_3d} we can see the mean squared error for both velocity components and temperature stabilizes on the order of $\mathcal{O}(10^{-4})$ for velocity components and $\mathcal{O}(10^{-1})$ for temperature). This stabilization is a crucial indicator of the long-term reliability of the hybrid approach in a higher-dimensional setting.

Beyond quantitative metrics, a method's ability to reproduce complex, large-scale flow structures is critical for its adoption in scientific discovery. A visual comparison of the intricate 3D flow field at a late time point (t = 30s) confirms a striking qualitative agreement between the hybrid prediction (\textbf{Fig.~\ref{fig:results_fig4}(c)}) and the OpenFOAM baseline (\textbf{Fig.~\ref{fig:results_fig4}(d)}). The intricate patterns of the streamlines and the overall flow topology are faithfully captured, providing intuitive and powerful evidence of the method's physical fidelity.

\begin{figure*}[p]
\centering
\includegraphics[width=0.8\textwidth]{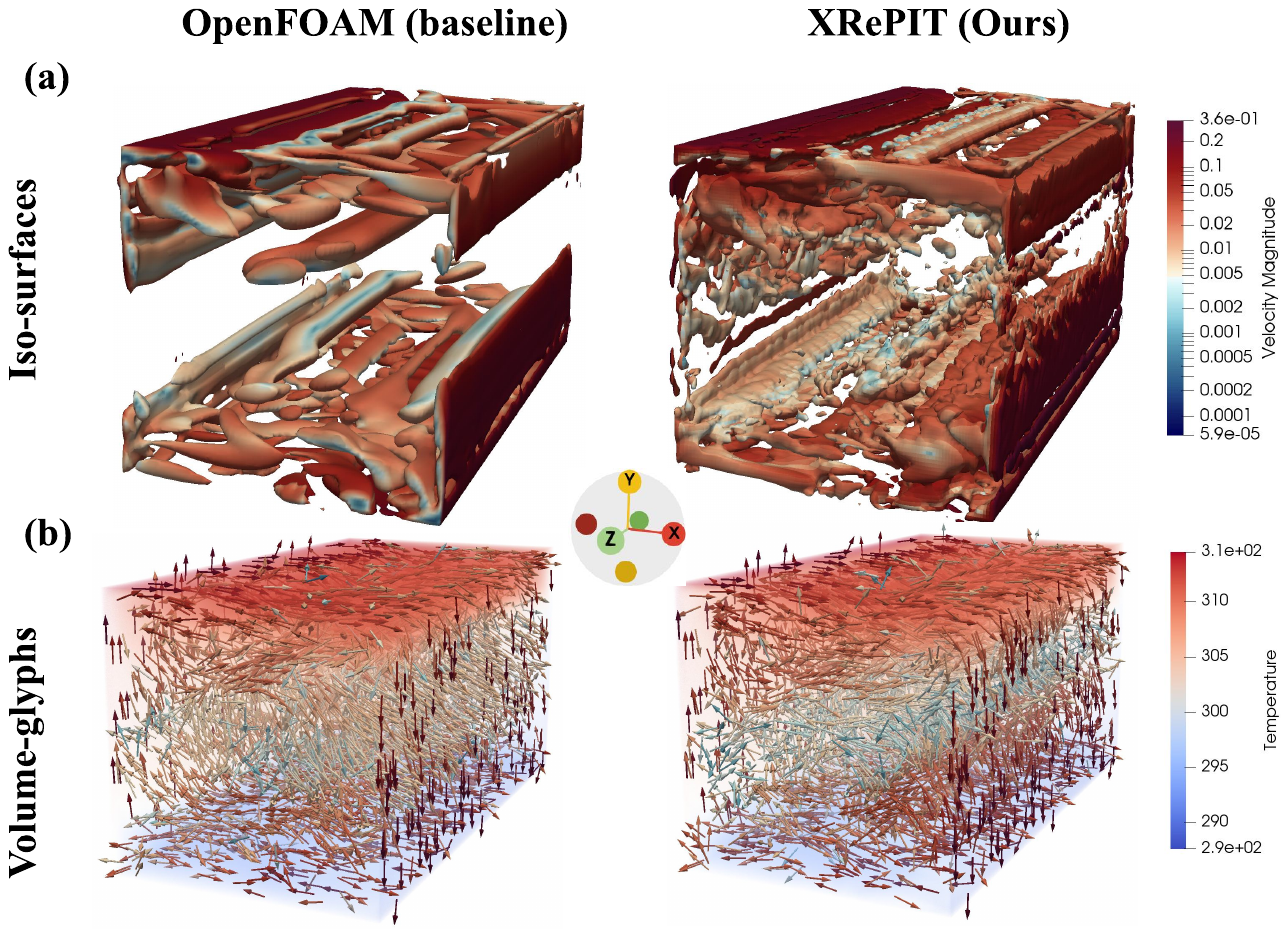}
\caption{\textbf{3D structural fidelity via Q-criterion iso-surfaces and thermal volume-glyphs at t = 30s.}
Left: ground truth (OpenFOAM). Right: XRePIT (ours).
\textbf{(a)} Q-criterion iso-surfaces at $Q=0.05$, highlighting rotation-dominated vortical cores.
\textbf{(b)} Temperature field with volume-glyph velocity vectors, illustrating agreement in macro-scale plume structure and primary recirculation while revealing small-scale morphological differences near the resolution limit.}
\label{fig:iso_surfaces_3D}
\end{figure*}

The volumetric rendering of the temperature field shown in \textbf{Fig.~\ref{fig:iso_surfaces_3D}(b)}, overlaid with velocity vectors, confirms that XRePIT accurately reconstructs the dominant macro-scale physics. The hybrid solver captures the characteristic mushroom-shaped thermal plume rising from the heated wall, properly reproducing the stratification layers near the adiabatic ceiling. The velocity glyphs also show strong alignment in both magnitude and direction within the primary recirculation zones. This indicates that the momentum transfer (driven by the buoyancy term in the Navier-Stokes equations) is being correctly sustained by the surrogate even after thousands of autoregressive steps.

To scrutinize the capture of shear layers, we computed the Q-criterion isosurfaces ($Q=0.05$) as seen in \textbf{Fig.~\ref{fig:iso_surfaces_3D}(a)}. The Q-criterion identifies regions where the magnitude of the rotation rate dominates the strain rate, effectively isolating coherent vortex cores. While both solutions exhibit vortical tubes forming along the shear boundaries, a notable morphological discrepancy is observed: the XRePIT solution displays a higher density of fragmented, small-scale isosurfaces compared to the smoother ground truth. This difference highlights the distinct numerical signatures of the two methods. The Finite Volume solver employs diffusive operators (viscosity and numerical discretization schemes) which naturally smooth out discontinuities, resulting in coherent, continuous tubes. In contrast, while deep learning models typically exhibit spectral bias, favoring the learning of lower-frequency (macro) features, the fragmentation observed here is indicative of high-frequency prediction noise. Lacking the explicit smoothness constraints of the differential operator, the surrogate introduces minor spatial discontinuities or ``jitter'' at the shear interfaces. While the macro-topology (location and orientation of the vortex cores) is consistent with the reference, these fragmented artifacts suggest that the AI method preserves the bulk flow accurately but struggles to replicate the perfectly smooth gradients of the diffusive physics at the sub-grid scale.

This successful extension to 3D, achieving a 2.91× speedup (\textbf{Fig.~\ref{fig:acceleration_barplot}}) over 2,000 timesteps, demonstrates the practical scalability of the timestep-coupled hybrid approach. Among the studies surveyed in Table 1, this work is, to the best of our knowledge, the first to demonstrate an automated, residual-guided hybrid ML-CFD pipeline that maintains stable long-horizon 3D rollouts with bounded error growth. This result elevates the methodology from an academic concept to a scalable strategy for accelerating high-fidelity simulations in real-world scientific and engineering domains.

\section{Discussion}
\label{sec:discussion}

\subsection{Stability and adaptability of residual-guided hybridization}

Across thousands of timesteps, the hybrid trajectories remain bounded and recover from periods of surrogate drift by invoking CFD correction when the monitored residual indicates a loss of consistency. This behavior aligns with the intended role of residual-guided hybridization: not to eliminate numerical simulation, but to reduce its frequency while preserving stability. Throughout this article, robustness is defined operationally as maintaining bounded error metrics across all field variables over thousands of timesteps. Concurrently, the evidence supports a specific notion of ``generalization'': the model adapts online to new thermal boundary conditions while the PDE, solver family, mesh topology, and geometry are held fixed. This represents a meaningful distribution shift for design-space sweeps varying operating conditions, but it is not a claim of portability to new geometries or fundamentally different physics without retraining or architectural modification.

\subsection{Modularity and extensibility of the framework}

While hybrid AI-CFD simulations have appeared in prior studies, the primary contribution of XRePIT is their systematization into an automated, end-to-end pipeline serving as a benchmarking substrate. In particular, the coupling and transfer-learning logic can accommodate different surrogate backbones with minimal disruption, enabling controlled comparisons between architectures under an identical CFD interface. The 3D extension is critical in this context: it demonstrates that the coupling mechanism, training loop, and runtime orchestration are not inherently limited to low-dimensional prototypes, even if the present physics and geometry remain similar.

\subsection{Interaction of hybrid acceleration with parallel CFD execution}

Although the present study uses a single-core CFD baseline, the framework supports parallel execution. We analyze the interaction between hybrid acceleration and parallel CFD through two perspectives: computational cost and coupling overhead.

\paragraph{Computational perspective.}
Hybrid speedup (Eq.~\ref{eq:acceleration}) depends on the ratio of $t_{CFD}$ to $t_{ML}$. CFD cost scales with cell count ($N_c$) and inner iterations ($N_{\text{iter}}$), with parallel runs requiring MPI boundary-data exchanges and global reductions per iteration. Conversely, the ML surrogate replaces the iterative solver with a single forward pass, entirely independent of $N_{\text{iter}}$. Consequently, single-pass GPU inference consistently outperforms communication-bound parallel CFD wherever $N_{\text{iter}}$ is non-trivial. Table~\ref{tab:scaling_regimes} summarizes these dynamics.

\begin{table}[h]
\centering
\caption{Expected hybrid acceleration ($\psi$) and CFD parallel efficiency across computational regimes.}
\label{tab:scaling_regimes}
\small
\begin{tabular}{@{} p{2.0cm} p{3.2cm} p{2.0cm} p{6.2cm} @{}}
\toprule
\textbf{Regime} & \textbf{CFD parallel efficiency} & \textbf{Hybrid $\psi$} & \textbf{Rationale} \\
\midrule
Low $N_c$, low $N_{\text{iter}}$ & 
Poor &
Moderate &
Per-core workload too small for efficient domain decomposition. Both $t_{CFD}$ and $t_{ML}$ are small. \\
\addlinespace
Low $N_c$, high $N_{\text{iter}}$ &
Moderate &
Good &
$t_{CFD}$ grows with iterations; $t_{ML}$ stays constant. Surrogate bypasses the entire iterative sequence. \\
\addlinespace
High $N_c$, low $N_{\text{iter}}$ &
Good &
Good &
CFD parallelizes efficiently. ML scales via GPU parallelism without inter-process communication. \textbf{Present study ($N_c{=}2{\times}10^6$, $N_{\text{iter}}{=}2$).} \\
\addlinespace
High $N_c$, high $N_{\text{iter}}$ &
Good &
\textbf{Best} &
CFD scales well per iteration but pays $N_{\text{iter}}$ synchronization barriers per timestep. Surrogate replaces all with one forward pass. \\
\bottomrule
\end{tabular}
\end{table}

\paragraph{Pre/post-processing perspective.}
Unlike standalone continuous CFD, the hybrid workflow introduces frequent I/O overheads: decomposing ML-predicted fields (decomposePar) before each CFD burst, and reconstructing solver outputs (reconstructPar) for ML ingestion. These disk- and network-bound operations occur at every model switch. The serial baseline avoids these overheads entirely. The tested configuration ($N_c = 2 \times 10^6$, PIMPLE with 1 outer and 2 inner loops) falls into the high-$N_c$/low-$N_{\text{iter}}$ regime, where standalone parallel CFD would scale well for long uninterrupted runs. However, within the hybrid loop, the solver operates in short correction bursts (${\sim}10$ timesteps), for which the per-invocation MPI startup cost cannot be amortized, and the coupling I/O described above adds further overhead. The single-core baseline therefore represents the most conservative reference: it yields the fastest per-timestep CFD execution under these conditions, and any reported acceleration is measured against this strongest possible baseline. Systematic benchmarking against multi-core baselines on larger, stiffer problems is an important direction for future work. 

Strong and weak scaling experiments support this (Appendix~\ref{sup_sec:scaling}). Strong scaling tests on the $100 \times 100 \times 200$ mesh confirm serial execution remains the fastest per-timestep option for short bursts (Appendix Table~\ref{sup_tab:strong_scaling}). Furthermore, weak scaling tests show the hybrid speedup $\psi$ remains robust at $3$--$4\times$ across a 27-fold increase in problem size (Appendix Table~\ref{sup_tab:weak_scaling}). Detailed analysis of both experiments is provided in Appendix~\ref{sup_sec:scaling}.

\subsection{Limitations and future challenges}

These are the current limitations of this study and, thus, the future challenges to be tackled:
\begin{itemize}
    \item \emph{Geometry and PDE scope.} All demonstrations are performed on a single canonical configuration (natural convection) with a fixed mesh/geometry. As a result, the present evidence supports robustness \emph{within} that problem family but does not establish geometry-invariant generalization. Extending XRePIT to new geometries will require either retraining on the new mesh/representation or adopting geometry-aware architectures capable of accommodating changing domains.

    \item \emph{Flow-regime coverage.} The study does not include turbulence models, strongly advection-dominated regimes, shocks, multiphase interfaces, or sharp-gradient transport—precisely the cases where error growth mechanisms and residual signals may behave qualitatively differently. The long-horizon stability demonstrated here is an empirical result for the laminar/transitional buoyancy-driven regime, not a guarantee for higher Reynolds numbers or chaotic dynamics.

    \item \emph{Residual trigger sufficiency.} The switching logic relies on a mass-conservation residual. While necessary for incompressible consistency, this criterion is not sufficient for overall accuracy; in other regimes, momentum/energy imbalance or pressure--velocity coupling errors may dominate even when $\nabla\!\cdot\!u$ remains small. This is not a flaw in the hybrid concept, but rather the most direct lever that must be strengthened for broader reliability.
\end{itemize}

\section{Conclusion}
\label{sec:conclusion}
This work evaluates a residual-guided, timestep-coupled hybrid simulation strategy by developing a fully-automated cross-computation framework. Within the scope studied (buoyancy-driven weakly compressible flow), XRePIT demonstrates that (i) long-horizon rollouts can be stabilized through intermittent numerical correction, (ii) online transfer learning can adapt a surrogate to unseen boundary-condition settings, and (iii) the same coupling logic can be carried from 2D to 3D with measurable end-to-end acceleration. The central takeaway is that tightly integrating residual monitoring, switching, and lightweight online updates into a single workflow renders ML--CFD hybridization operationally reliable for long unsteady runs.

Finally, we view open-sourcing XRePIT as an integral part of the scientific contribution. The most credible path toward maturing hybrid ML--CFD methods lies in reproducible baselines, ablation-ready implementations, and community-driven validation across solvers, meshes, and regimes. By positioning XRePIT as an extensible engineering workflow rather than a universal solution, we aim to facilitate that maturation process and provide a practical reference point for future hybrid designs.

\section{Acknowledgements}
This work was supported by the Nuclear Safety Research Program through the Regulatory Research Management Agency for SMRs (RMAS) and the Nuclear Safety and Security Commission (NSSC) of the Republic of Korea. (No. RS-2024-00509653) and the National Research Council of Science \& Technology (NST) grant by the Korea government (MIST) (No. GTL24031-000).

 \section{Data availability and code release}
Data generation is part of the framework's processes. And, the source code for the proposed framework, along with a comprehensive README and usage instructions, will be made publicly available on GitHub at: \url{https://github.com/POSTECH-NINE/repitframework}. The repository will be open to the public upon the publication of this manuscript in a peer-reviewed journal. Users may request support or report issues through the repository’s issue tracker.

\section{CRediT authorship contribution statement}
\textbf{Shilaj Baral:} Writing -- original draft, Validation, Software, Methodology, Investigation. \textbf{Youngkyu Lee:} Writing -- review/editing, Methodology, Investigation, Software. \textbf{Sangam Khanal:} Writing -- review/editing, Software.  \textbf{Joongoo Jeon:} Writing -- review/editing, Validation, Methodology, Investigation, Supervision, Conceptualization, Funding acquisition.

\section{Competing interests}
The authors declare that they have no known competing financial interests or personal relationships that could have appeared to influence the work reported in this paper.

\bibliographystyle{elsarticle-num}
\bibliography{references}

\appendix
 \setcounter{figure}{0}
 \setcounter{table}{0}
 \setcounter{equation}{0}
\renewcommand{\thefigure}{A\arabic{figure}}
\renewcommand{\tablename}{Appendix Table}
\renewcommand{\thetable}{A\arabic{table}}
\renewcommand{\theequation}{A\arabic{equation}}

\section{Neural network information}
This section describes the neural network architectures used in this study and lists the full hyperparameter configurations required for reproducibility.

\subsection{Targeted improvements in the FVMN}
\label{sup_subsec:fvmn}
To further optimize the FVMN within the XRePIT workflow, we introduced several targeted improvements. First, Optuna~\cite{akiba2019optuna} was used to tune the model architecture, yielding a configuration with three hidden layers of width 398 and a learning rate of 0.001. During transfer learning, freezing the first layer consistently improved results; this strategy was adopted throughout. Additionally, batch normalization~\cite{ioffe2015batchNorm} was applied after each hidden layer (but not after the output layer), and dropout~\cite{srivastava2014dropout} was applied after the last hidden layer to reduce overfitting and enhance generalization. Data pre-/post-processing routines were streamlined relative to the baseline implementation, improving end-to-end efficiency. Apart from these changes, the original architecture was left unchanged. A parametric overview is provided in Appendix Table~\ref{sup_tab:fvmn_parameters}.

\begin{table}[htbp]
  \centering
  \caption{%
    Hyperparameters for the extended FVMN architecture.
    Each flow variable is modeled by an independent network with this configuration,
    trained on a combined loss.}
  \label{sup_tab:fvmn_parameters}

  \begin{threeparttable}
  \footnotesize
  \setlength{\tabcolsep}{6pt}
  \renewcommand{\arraystretch}{1.1}
  \begin{tabular}{@{}ll@{}}
    \toprule
    \textbf{Hyperparameter} & \textbf{Value} \\
    \midrule
    Linear layers & 5 (3 hidden) \\
    Layer width & 398 \\
    Optimizer & Adam \\
    Loss function & MSE \\
    Learning rate & 0.001 \\
    Batch-normalization layers & 4 \\
    Dropout (last hidden layer) & 0.2 \\
    Input features (per variable) & 15 (5 stencil points $\times$ 3 variables) \\
    Output features (per variable) & 1 (derivative) \\
    \bottomrule
  \end{tabular}
  \end{threeparttable}
\end{table}

\subsection{Derivation and hyperparameters for FVFNO}
\label{sup_subsec:fvfno}
The FVFNO architecture uses a Fourier Neural Operator (FNO) as its core processing block. FNOs learn mappings between function spaces via global convolutions performed in the frequency domain~\cite{li2020FNO}.

We cast the learning problem as operator approximation. Given input–output pairs of functions $(\zeta, d\zeta)$ supported on a 2D spatial grid $\Omega$, the goal is to approximate a solution operator $\mathcal{G}^\dagger\!: Z \to dZ$ mapping an input $\zeta \in Z$ to its associated output $d\zeta \in dZ$ (time derivatives) as defined in Eqs.~\eqref{eq:tiered_input} and~\eqref{eq:derivative_output}:
\begin{equation}
  \label{eq:tiered_input}
  \zeta^{t}(x) = \big[\zeta_{i,j}^{t},\, \zeta_{i-1,j}^{t},\, \zeta_{i+1,j}^{t},\, \zeta_{i,j-1}^{t},\, \zeta_{i,j+1}^{t}\big],
\end{equation}
where $\zeta_{i,j}^{t}$ denotes the field snapshot at spatial index $(i,j)\in\Omega$ and time $t$.
\begin{equation}
  \label{eq:derivative_output}
  (d\zeta)^{t}(x) = \zeta_{i,j}^{t+1} - \zeta_{i,j}^{t}.
\end{equation}
Given a dataset $\{(\zeta_k, d\zeta_k)\}_{k=1}^{N}$ with $\zeta_k \sim \mu$ i.i.d.\ on $Z$, we learn a parameterized operator $\mathcal{G}_{\theta}\!: Z \to dZ$ with $\theta \in \Theta$ by minimizing a suitable cost $C$ (e.g., MSE):
\begin{equation}
  \min_{\theta\in\Theta}\;
  \mathbb{E}_{\zeta\sim \mu}\!\left[\,C\!\big(\mathcal{G}_{\theta}(\zeta),\, \mathcal{G}^{\dagger}(\zeta)\big)\,\right].
\end{equation}

Let $\zeta \in \mathbb{R}^{b\times i}$ (with $b$ grid points or batch size and $i$ input features). We arrange $\zeta$ as $\mathbb{R}^{b\times i\times 1}$ and lift it to a channel width $w$ via a linear map

\begin{equation}
  h_{0} = \mathcal{L}\,\zeta, \qquad
  \mathcal{L}: \mathbb{R}^{b\times i\times 1}\to \mathbb{R}^{b\times i\times w}.
\end{equation}

Rearranging to $h_{0}\in \mathbb{R}^{b\times w\times i}$, we apply $L$ Fourier layers ($\ell=0,\dots,L-1$), each performing a global spectral convolution plus a local transform:

\begin{equation}
  h_{\ell+1} = \sigma\!\Big(\,\mathcal{F}^{-1}\!\big[\mathcal{F}(h_{\ell}) \odot R^{(\ell)}\big] + W^{(\ell)} h_{\ell}\Big),
\end{equation}

 where: 
 \begin{itemize}
     \item $\mathcal{F}$ and $\mathcal{F}^{-1}$ are the (discrete) Fourier and inverse Fourier transforms.
     \item $R^{(l)}$ is learnable complex tensor (the Fourier-space kernel), truncated to the lowest $m$ modes.
     \item $\odot$ denotes mode-wise matrix multiplication:
        \begin{equation}
            \bigg[ \mathcal{F}(h_l) \odot R^{(l)} \bigg]_{b,w,m} = \sum_{i=1}^w  \mathcal{F}(h_l)_{b,i,m} R^{(l)}_{i,w,m}
        \end{equation}
    \item $W^{(\ell)}$ is a learnable pointwise (local) operator.
    \item $\sigma$ is a nonlinearity (with optional BatchNorm and Dropout).
 \end{itemize}

Finally, linear projections produce $o$ output features:
\begin{equation}
  d\zeta = \mathcal{P}_2\!\big(\mathcal{P}_1(h_{L})\big),
\end{equation}
with $\mathcal{P}_1:\mathbb{R}^{b\times w\times i}\to \mathbb{R}^{b\times (w i)}$ (flatten) and
$\mathcal{P}_2:\mathbb{R}^{b\times (w i)}\to \mathbb{R}^{b\times o}$ (linear).
The network outputs $d\zeta$, which is added to the current state to obtain the next predicted field. The key hyperparameters appear in Appendix Table~\ref{sup_tab:fvfno_parameters}.

\begin{table}[htbp]
  \centering
  \caption{%
    Hyperparameters for the FVFNO architecture.
    Each flow variable is modeled by an independent network with this configuration,
    trained on a combined loss.}
  \label{sup_tab:fvfno_parameters}

  \begin{threeparttable}
  \footnotesize
  \setlength{\tabcolsep}{6pt}
  \renewcommand{\arraystretch}{1.1}
  \begin{tabular}{@{}ll@{}}
    \toprule
    \textbf{Hyperparameter} & \textbf{Value} \\
    \midrule
    Fourier layers & 3 \\
    Frequency modes & 12 \\
    Layer width & 64 \\
    Optimizer & Adam \\
    Loss function & MSE \\
    Learning rate & 0.001 \\
    Activation function & ReLU \\
    Batch-normalization layers & 5 \\
    Dropout (last linear layer) & 0.2 \\
    Input features (per variable) & 15 (5 stencil points $\times$ 3 variables) \\
    Output features (per variable) & 1 (derivative) \\
    \bottomrule
  \end{tabular}
  \end{threeparttable}
\end{table}

\subsection{Architectural benchmarking: extended error analysis}
\label{sup_subsec:error_analysis_networks}
\begin{figure*}[htbp]
  \centering
  \includegraphics[width=\textwidth]{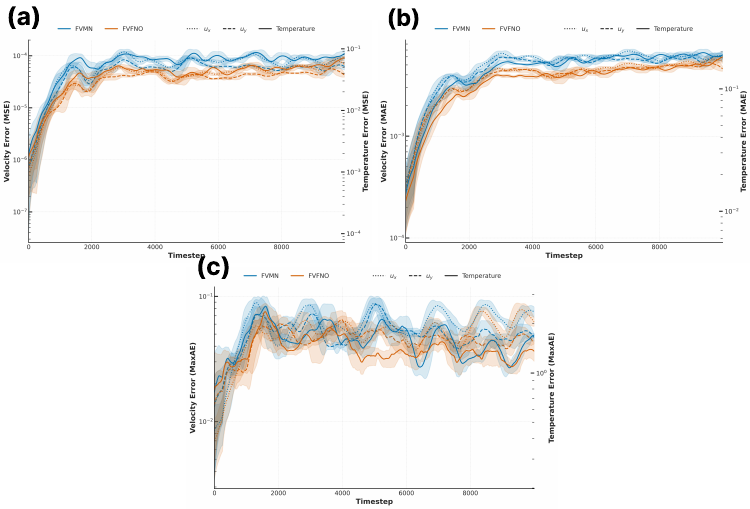}
  \caption{\textbf{Comparative error analysis of neural network architectures.}
  Performance of \textbf{FVMN} (blue) and \textbf{FVFNO} (orange) within the hybrid framework over a 10{,}000-timestep rollout.
  A dual-axis plot shows \textbf{(a)} MSE, \textbf{(b)} MAE, and \textbf{(c)} MaxAE, with velocity errors ($u_x$, $u_y$) on the left axis and temperature ($T$) on the right.
  The framework maintains stable, non-divergent error profiles for both models, with \textbf{FVFNO} consistently achieving lower errors across all fields.}
  \label{sup_fig:errors_network_comparison}
\end{figure*}

To supplement the analysis in the main text, Fig.~\ref{sup_fig:errors_network_comparison} compares Mean Squared Error (MSE), Mean Absolute Error (MAE), and Maximum Absolute Error (MaxAE) for each physical field when using FVMN and FVFNO in the hybrid loop.

\section{Grid convergence study for 3D natural circulation}
\label{sec:grid_convergence}
A systematic grid convergence study was performed following the procedure recommended by Celik et al. ~\cite{celik2008procedure} to quantify the spatial discretization uncertainty in the numerical results. Three progressively refined hexahedral meshes were employed, as summarised in Table~\ref{tab:grid_params}.

\begin{table}[htbp]
  \centering
  \caption{Grid parameters for the convergence study.}
  \label{tab:grid_params}
  \begin{tabular}{lccc}
    \hline
    Parameter & Coarse & Medium & Fine \\
    \hline
    Grid ($N_x \times N_y \times N_z$) & $65\times65\times130$ & $100\times100\times200$ & $135\times135\times270$ \\
    Total cells, $N$                   & 549\,250              & 2\,000\,000             & 4\,920\,750 \\
    Representative size, $h$ (m)       & 0.01154               & 0.00750                 & 0.00556 \\
    \hline
  \end{tabular}
\end{table}

The representative grid size for each mesh is defined as
\begin{equation}\label{eq:grid_size}
  h = \left(\frac{V_{\text{domain}}}{N}\right)^{1/3},
\end{equation}
where $V_{\text{domain}} = 0.75 \times 0.75 \times 1.5 = 0.844\;\text{m}^3$ is the computational domain volume and $N$ is the total number of cells. The refinement ratios between successive grids are $r_{21} = h_{\text{coarse}}/h_{\text{mid}} = 1.538$ and $r_{32} = h_{\text{mid}}/h_{\text{fine}} = 1.350$.

The key variable selected for the convergence study is the time-averaged Nusselt number on the hot wall, $\overline{\text{Nu}}_h$, computed at the mid-depth plane ($Z = 0$) in order to compare with the experimental results defined in ~\cite{ampofo2003experimental}. Since the refinement ratios are non-uniform ($r_{21} \neq r_{32}$), the apparent order of convergence $p$ is obtained by solving the following expression iteratively:
\begin{equation}\label{eq:apparent_order}
  p = \frac{1}{\ln(r_{21})} \left| \ln\left|\frac{\varepsilon_{32}}{\varepsilon_{21}}\right| + \ln\!\left(\frac{r_{21}^{p} - s}{r_{32}^{p} - s}\right) \right|,
\end{equation}
where $\varepsilon_{32} = \phi_3 - \phi_2$, $\varepsilon_{21} = \phi_2 - \phi_1$, $s = \text{sign}(\varepsilon_{32}/\varepsilon_{21})$, and the subscripts 1, 2, 3 denote the fine, medium, and coarse grids, respectively. The Richardson-extrapolated value is then calculated as
\begin{equation}\label{eq:richardson}
  \phi_{\text{ext}}^{21} = \frac{r_{21}^{p}\,\phi_1 - \phi_2}{r_{21}^{p} - 1}.
\end{equation}

The approximate relative error, the extrapolated relative error, and the Grid Convergence Index with a safety factor $F_s = 1.25$ are defined as
\begin{equation}\label{eq:errors}
  e_a^{21} = \left|\frac{\phi_1 - \phi_2}{\phi_1}\right|, \qquad
  e_{\text{ext}}^{21} = \left|\frac{\phi_{\text{ext}}^{21} - \phi_1}{\phi_{\text{ext}}^{21}}\right|, \qquad
  \text{GCI}_{\text{fine}}^{21} = \frac{F_s \, e_a^{21}}{r_{21}^{p} - 1}.
\end{equation}

The results of the convergence study are presented in Table~\ref{tab:gci_results}.

\begin{table}[htbp]
  \centering
  \caption{Grid Convergence Index results for $\overline{\text{Nu}}_h$.}
  \label{tab:gci_results}
  \begin{tabular}{lc}
    \hline
    Quantity & Value \\
    \hline
    $\phi_1$ (fine grid)                            & 57.497 \\
    $\phi_2$ (medium grid)                          & 55.366 \\
    $\phi_3$ (coarse grid)                          & 45.976 \\
    Apparent order of convergence, $p$              & 5.340 \\
    Extrapolated value, $\phi_{\text{ext}}^{21}$    & 57.734 \\
    Approximate relative error, $e_a^{21}$          & 3.705\% \\
    Extrapolated relative error, $e_{\text{ext}}^{21}$ & 0.411\% \\
    $\text{GCI}_{\text{fine}}^{21}$                 & 0.516\% \\
    $\text{GCI}_{\text{mid}}^{32}$                  & 5.345\% \\
    Asymptotic range indicator                      & 1.039 \\
    \hline
  \end{tabular}
\end{table}

The fine grid solution ($\overline{\text{Nu}}_h = 57.50$) lies within 0.41\% of the Richardson-extrapolated estimate of 57.73, and the corresponding $\text{GCI}_{\text{fine}}^{21}$ of 0.52\% indicates a low level of numerical uncertainty. The asymptotic range indicator of 1.039 is close to unity, confirming that the three grids are within the asymptotic convergence regime. Although the fine grid yields marginally closer agreement with the 
Richardson-extrapolated estimate, the medium grid 
($100\times100\times200$) already captures the hot-wall Nusselt number 
profile to within 3.71\% of the fine grid solution, and its 
$\text{GCI}_{\text{mid}}^{32}$ of 5.35\% is well within commonly accepted 
engineering thresholds. As shown in Fig.~\ref{fig:grid_convergence}, the 
medium grid reproduces the key features of the experimental Nusselt 
number distribution from~\cite{ampofo2003experimental}---including the 
peak values near the cavity top and bottom and the relatively uniform 
mid-height region---with no qualitative difference from the fine grid. 
Given that the fine mesh contains approximately 2.5 times more cells 
(4.92\,M vs.\ 2\,M), resulting in a proportionally higher computational 
cost per timestep, and that the primary objective of this study is to 
benchmark the hybrid ML--CFD coupling rather than to pursue 
grid-independent DNS-level accuracy, the medium grid provides an 
appropriate balance between spatial fidelity and computational 
tractability. It is therefore adopted for all subsequent 3D analyses 
presented in this work.
\section{Strong and weak scaling analysis}
\label{sup_sec:scaling}
 
To empirically characterize the interaction between hybrid acceleration and parallel CFD execution, we conducted two scaling experiments using the XRePIT framework on an AMD EPYC 9554 (256-core) CPU with an NVIDIA A100 (40\,GB) GPU. All runs use PIMPLE with 1 outer and 2 inner correction loops and a residual threshold of 5.
 
\paragraph{Strong scaling.}
Table~\ref{sup_tab:strong_scaling} reports metrics for the present $100 \times 100 \times 200$ mesh ($N_c = 2 \times 10^6$) with $n_p \in \{1, 2, 5, 10\}$ processors. The measured $t_{CFD}$ includes both the solver execution and post-burst reconstruction (\texttt{reconstructPar -newTimes}), since both are required before the surrogate can resume. Serial execution ($n_p = 1$) yields the lowest per-timestep CFD cost because it avoids MPI initialization, inter-process communication, and file-based reconstruction entirely. Adding processors increases $t_{CFD}$ by ${\sim}7\times$ due to the dominance of these overheads over the marginal compute savings on short 10-timestep correction bursts. The reported $\psi$ for $n_p > 1$ should be interpreted with care: it compares hybrid time against a hypothetical all-parallel CFD baseline, both of which carry the same reconstruction overhead. The serial $\psi = 2.88$ represents the most conservative and physically meaningful acceleration.
 
\begin{table}[h]
\centering
\caption{Strong scaling analysis. Fixed mesh: $100 \times 100 \times 200$ ($N_c = 2 \times 10^6$). $t_{CFD}$ includes solver execution and reconstruction. $\psi$ is computed relative to a CFD-only baseline at the same $n_p$.}
\label{sup_tab:strong_scaling}
\small
\begin{tabular}{@{} r r r r r r r @{}}
\toprule
$n_p$ & $t_{CFD}$ (s) & $t_{ML}$ (s) & $t_{up}$ (s) & CFD-only (s) & XRePIT (s) & $\psi$ \\
\midrule
1  & 22.8   & 1.712 & 10.0  & 22{,}900  & 7{,}933  & 2.88 \\
2  & 166.78 & 1.668 & 9.78  & 167{,}610 & 36{,}655 & 4.57 \\
5  & 160.09 & 1.678 & 9.69  & 161{,}206 & 35{,}309 & 4.56 \\
10 & 157.71 & 1.612 & 9.61  & 158{,}812 & 34{,}764 & 4.56 \\
\bottomrule
\end{tabular}
\end{table}
 
\paragraph{Weak scaling.}
Table~\ref{sup_tab:weak_scaling} reports metrics for three mesh resolutions with a fixed workload of ${\sim}4{,}096$ cells per processor, following Gustafson's scaling model. As the problem size grows from 8{,}192 to 221{,}184 cells, $t_{CFD}$ increases from 0.203\,s to 2.795\,s per timestep, reflecting the growing per-processor workload and inter-partition communication. $t_{ML}$ scales proportionally with $N_c$ (from 0.008\,s to 0.176\,s), confirming that GPU inference cost grows with the forward-pass batch size but remains an order of magnitude below $t_{CFD}$ at all scales tested. The hybrid speedup $\psi$ remains in the $3$--$4\times$ range across a 27-fold increase in problem size. The slight decrease at the largest mesh ($\psi = 3.05$) is attributable to the growth of the per-switch update cost $t_{up}$, which increases from 0.014\,s to 0.769\,s as the online transfer-learning step processes larger fields.
 
\begin{table}[h]
\centering
\caption{Weak scaling analysis. Fixed workload: ${\sim}4{,}096$ cells/processor. PIMPLE: 1 outer, 2 inner corrections.}
\label{sup_tab:weak_scaling}
\small
\begin{tabular}{@{} l r r r r r r r r @{}}
\toprule
Grid & $N_c$ & $n_p$ & $t_{CFD}$ (s) & $t_{ML}$ (s) & $t_{up}$ (s) & CFD-only (s) & XRePIT (s) & $\psi$ \\
\midrule
$16 \times 16 \times 32$  & 8{,}192   & 2  & 0.203 & 0.008 & 0.014 & 204.9   & 55.8   & 3.67 \\
$32 \times 32 \times 64$  & 65{,}536  & 16 & 0.834 & 0.061 & 0.069 & 834.7   & 204.5  & 4.08 \\
$48 \times 48 \times 96$  & 221{,}184 & 54 & 2.795 & 0.176 & 0.769 & 2{,}814.6 & 922.4 & 3.05 \\
\bottomrule
\end{tabular}
\end{table}

\section{Figures and tables}

\begin{figure*}[htbp]
  \centering
  \includegraphics[height=0.8\textheight,keepaspectratio]{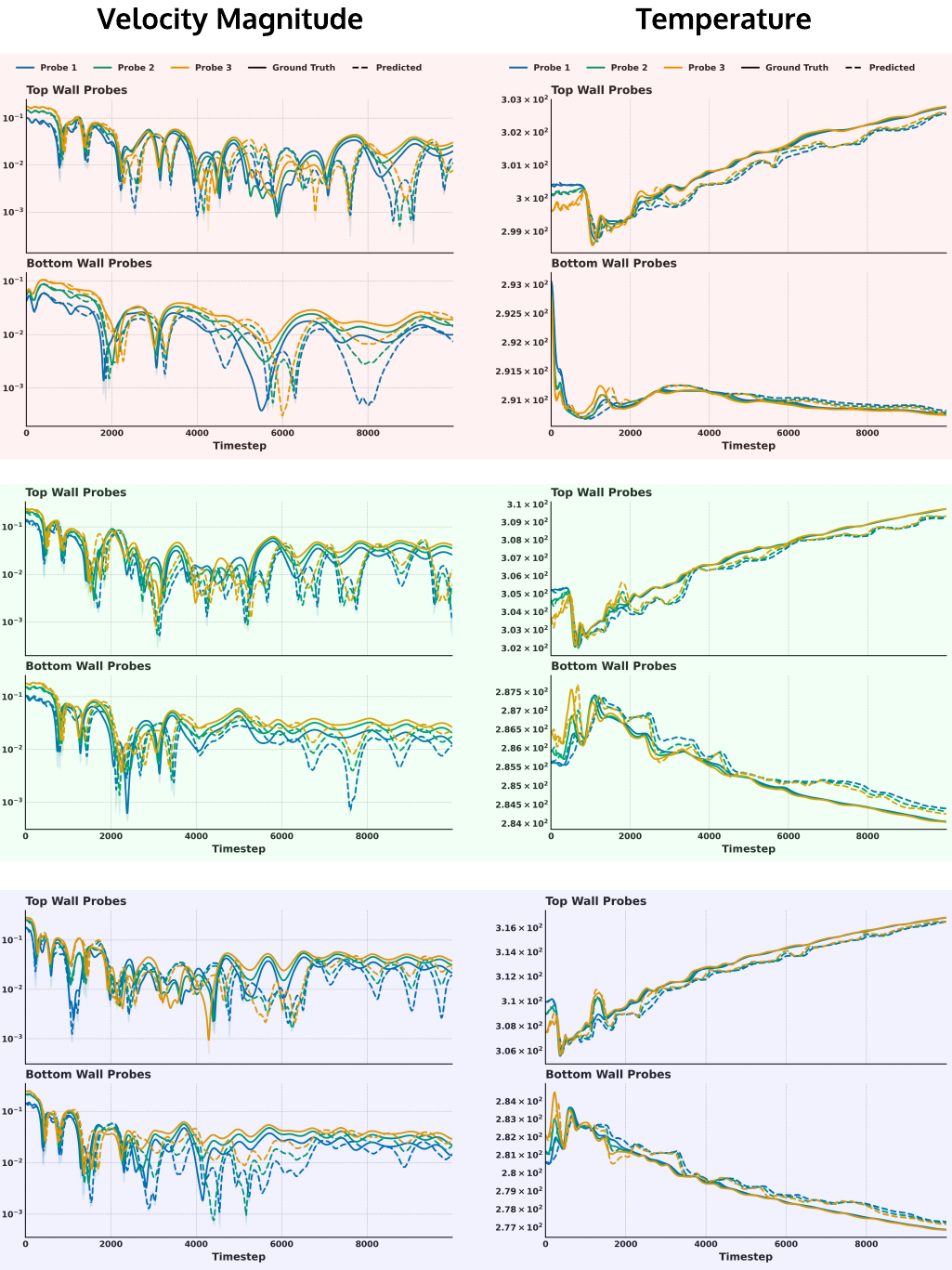}
  \caption{\textbf{Long-term stability and accuracy at local probe locations.}
  Temporal evolution of velocity magnitude ($|U|=\sqrt{u_x^{2}+u_y^{2}}$) and temperature ($T$) at six probe locations for three 2D generalization cases.
  Hybrid predictions (dashed) are compared with ground-truth CFD (solid).
  In \textbf{Case 1} (red), the model was initially trained for more epochs and then used for hybrid co-simulation; for \textbf{Case 2} (green) and \textbf{Case 3} (blue), the same pre-trained model from Case~1 was adapted via two epochs of transfer learning.
  The number of ML time steps per switch stabilizes as hybrid training proceeds.
  No error accumulation is observed in any case.}
  \label{sup_fig:probe_analysis_all}
\end{figure*}

\begin{figure*}[htbp]
  \centering
  \includegraphics[width=\textwidth]{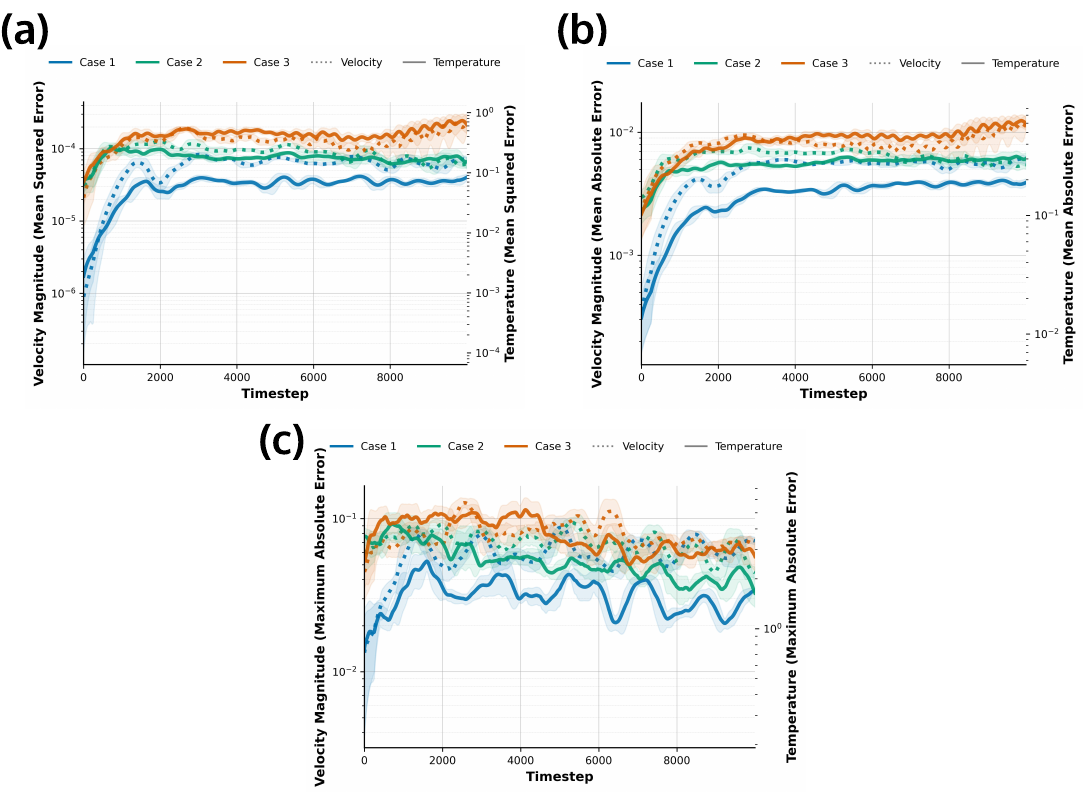}
  \caption{\textbf{Temporal evolution of domain-wide error metrics for the 2D generalization cases.}
  Comparison of \textbf{Case 1} (training condition) and two unseen generalization cases (\textbf{Case 2}, \textbf{Case 3}).
  The hybrid method maintains low, stable error profiles across boundary conditions, indicating robust generalization.
  \textbf{(a)} MSE for velocity magnitude and temperature.
  \textbf{(b)} MAE, showing similarly stable, low-magnitude errors.
  \textbf{(c)} MaxAE, confirming the absence of error divergence and long-term stability.}
  \label{sup_fig:error_analysis_cases}
\end{figure*}

\begin{figure*}[htbp]
  \centering
  \includegraphics[width=\textwidth]{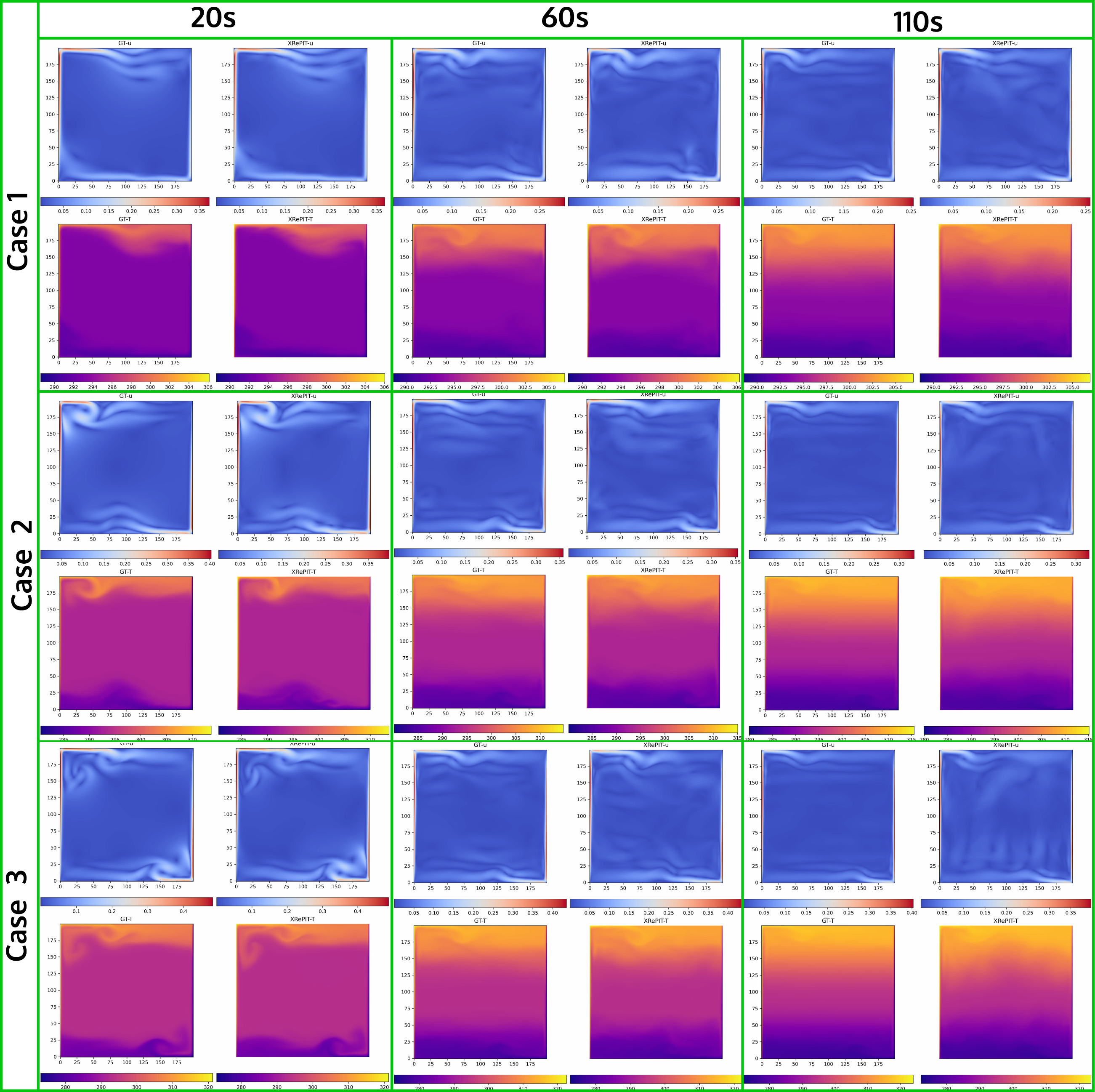}
  \caption{\textbf{High-fidelity validation of XRePIT across multiple boundary conditions.}
  Visual comparison of velocity magnitude ($U$) and temperature ($T$) predicted by XRePIT against ground truth (OpenFOAM).
  Rows correspond to \textbf{Case 1} (trained), \textbf{Case 2} (unseen), and \textbf{Case 3} (unseen).
  Columns show snapshots at 20\,s, 60\,s, and 110\,s, demonstrating long-term stability and accurate reconstruction of flow structures with rapid transfer learning.}
  \label{sup_fig:visual_comparison}
\end{figure*}

\begin{figure*}[htbp]
  \centering
  \includegraphics[width=\textwidth]{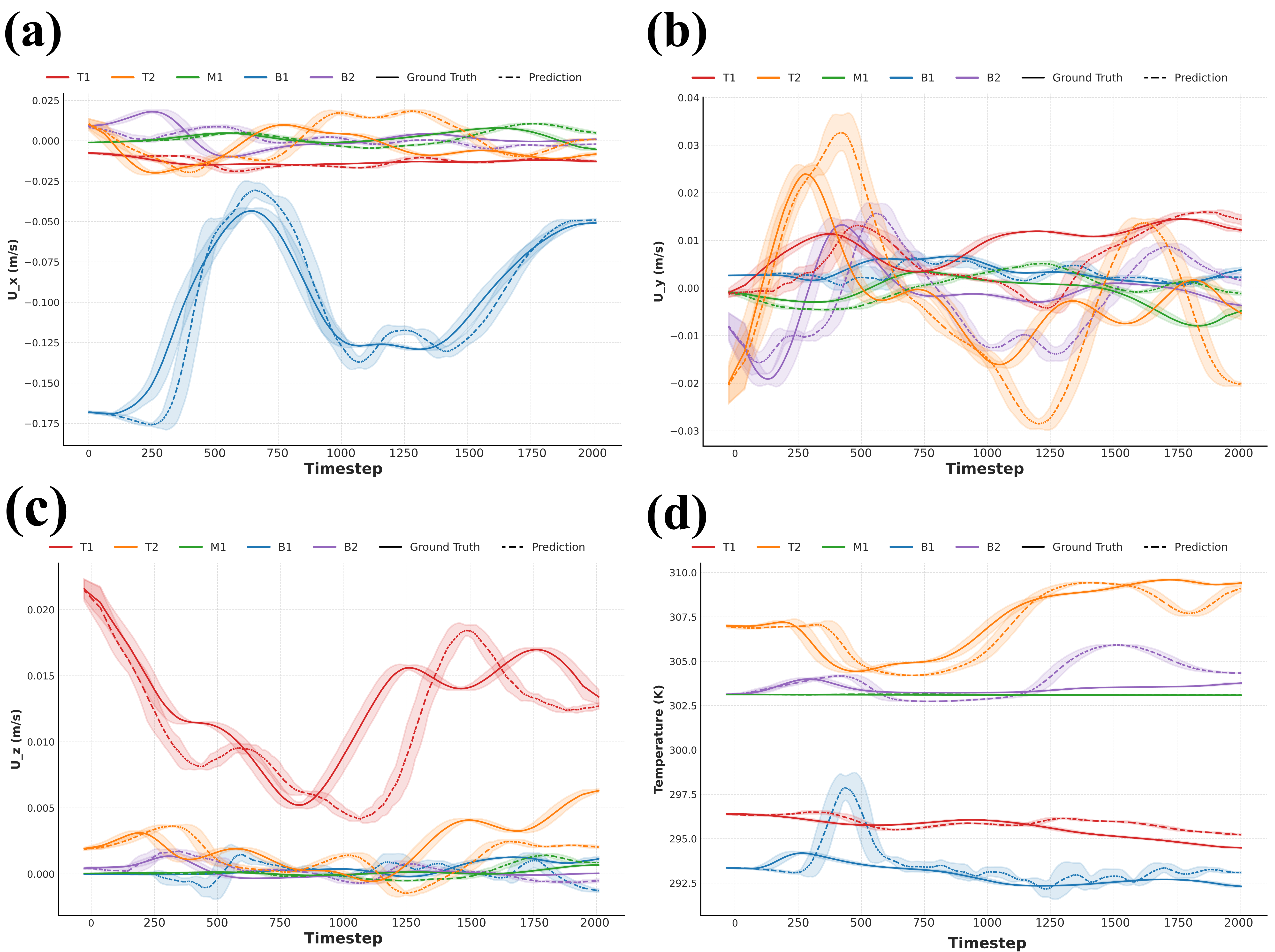}
  \caption{\textbf{Localized validation of long-term stability at 3D probe locations.}
  Point-wise comparison between hybrid predictions (dashed) and ground truth (solid) at five probe locations over a 2{,}000-timestep 3D simulation.
  \textbf{(a,b)} $u_x$ and $u_y$ closely track the primary dynamics with minor, bounded deviations.
  \textbf{(c)} $u_z$ is largely accurate; the \textbf{T2} probe (orange) exhibits an initial transient error that does not diverge due to intermediate physics-based corrections, likely reflecting the use of only two transfer-learning epochs in 3D.
  \textbf{(d)} Temperature ($T$) closely matches ground truth across probes.
  Overall, the framework maintains long-horizon stability with acceptable accuracy despite localized transients.}
  \label{sup_fig:probe_analysis_3d}
\end{figure*}

\begin{figure*}[htbp]
  \centering
  \includegraphics[width=\textwidth]{supplementary_figures/Figure_A6.png}
  \caption{\textbf{Quantitative error analysis in the 3D simulation.}
  Temporal evolution of whole-field prediction error over 2{,}000 timesteps.
  Each panel shows velocity component errors ($u_x,u_y,u_z$; left axis) and temperature error ($T$; right axis).
  \textbf{(a)} \textbf{MSE} for the velocity components remains on the order of $10^{-4}$ and stable; temperature MSE stays below $0.1$ without divergence.
  \textbf{(b)} \textbf{MAE} remains below $10^{-2}$ for velocities and stabilizes around $0.2$ for temperature.
  \textbf{(c)} \textbf{MaxAE} (worst-case deviation) is bounded across all fields, with higher initial values that later decrease and stabilize.}
  \label{sup_fig:error_analysis_3d}
\end{figure*}

\begin{sidewaystable*}[htbp]
  \centering
  \caption{%
    \textbf{Overall performance analysis for Case 2.}
    Relaxing the residual threshold hands off more aggressively to the ML surrogate, increasing prediction error in regimes with steeper thermal gradients.
    A stricter threshold with minimal retraining provides the best accuracy--acceleration trade-off.}
  \label{sup_tab:overall_performance_case2}

  \begin{threeparttable}
  \footnotesize

  \vspace{0.75ex}
  \textbf{(a) Acceleration analysis across epoch and residual-threshold configurations.}
  \vspace{0.5ex}

  \setlength{\tabcolsep}{3.5pt}
  \renewcommand{\arraystretch}{1.1}
  \begin{tabular}{@{}cccccccccccc@{}}
    \toprule
    \textbf{Epochs} & \textbf{Res.} &
    $\boldsymbol{t_{\mathrm{CFD}}}$ & $\boldsymbol{t_{\mathrm{ML}}}$ & $\boldsymbol{t_{\mathrm{up}}}$ &
    $\boldsymbol{n_{\mathrm{switch}}}$ & $\boldsymbol{n_{\mathrm{CFD}}}$ & $\boldsymbol{n_{\mathrm{ML}}}$ &
    \textbf{OpenFOAM (s)} & \textbf{XRePIT (s)} &
    $\boldsymbol{\psi}$ & $\boldsymbol{t_{\mathrm{avg.}}}$ \\
    \midrule
    2  &   5  & 0.74 & 0.026 & 1.24 & 365 & 3643 & 6364 & 7498.4 & 3354.4 & 2.23 & 17.43 \\
    2  &  10  & 0.75 & 0.026 & 1.28 & 307 & 3063 & 6937 & 7578.1 & 2898.3 & 2.61 & 22.59 \\
    2  & 100  & 0.73 & 0.026 & 1.31 & 178 & 1773 & 8227 & 7395.6 & 1761.5 & 4.19 & 46.21 \\
    10 &   5  & 0.75 & 0.025 & 6.24 & 331 & 3303 & 6697 & 7512.7 & 4720.3 & 1.59 & 20.23 \\
    10 &  10  & 0.75 & 0.026 & 6.41 & 289 & 2883 & 7117 & 7546.7 & 4215.5 & 1.79 & 24.62 \\
    10 & 100  & 0.73 & 0.026 & 6.57 & 176 & 1753 & 8247 & 7345.5 & 2665.1 & 2.75 & 46.85 \\
    \bottomrule
  \end{tabular}

  \vspace{1.25ex}
  \textbf{(b) Time-averaged spatial error metrics (Case 2).}
  \vspace{0.5ex}

  \setlength{\tabcolsep}{4pt}
  \renewcommand{\arraystretch}{1.1}
  \begin{tabular}{@{}cccccccccc@{}}
    \toprule
    \textbf{Epochs} & \textbf{Res.} &
    \textbf{L$_2$(T)} & \textbf{MSE(T)} & \textbf{MAE(T)} & \textbf{MaxAE(T)} &
    \textbf{L$_2$(U)} & \textbf{MSE(U)} & \textbf{MAE(U)} & \textbf{MaxAE(U)} \\
    \midrule
    2  &   5  & 1.50e$-$3 & 0.199 & 0.292 & 2.79 & 0.289 & 1.93e$-$4 & 0.010 & 0.082 \\
    2  &  10  & 1.93e$-$3 & 0.330 & 0.364 & 3.33 & 0.359 & 2.96e$-$4 & 0.013 & 0.091 \\
    2  & 100  & 4.11e$-$3 & 1.546 & 0.827 & 5.77 & 0.676 & 1.07e$-$3 & 0.026 & 0.154 \\
    10 &   5  & 1.17e$-$3 & 0.123 & 0.211 & 2.31 & 0.242 & 1.31e$-$4 & 0.009 & 0.068 \\
    10 &  10  & 1.50e$-$3 & 0.205 & 0.280 & 2.78 & 0.305 & 2.08e$-$4 & 0.011 & 0.079 \\
    10 & 100  & 3.90e$-$3 & 1.607 & 0.753 & 5.76 & 0.741 & 1.31e$-$3 & 0.027 & 0.147 \\
    \bottomrule
  \end{tabular}

  \end{threeparttable}
\end{sidewaystable*}

\begin{sidewaystable*}[htbp]
  \centering
  \caption{%
    \textbf{Case 3} exhibits sharper temperature drops and stronger buoyancy, increasing sensitivity to residual thresholds.
    Relaxing the threshold improves acceleration but degrades accuracy; conservative hand-off with minimal retraining gives the best balance.}
  \label{sup_tab:overall_performance_case3}

  \begin{threeparttable}
  \footnotesize

  \vspace{0.75ex}
  \textbf{(a) Acceleration analysis across epoch and residual-threshold configurations.}
  \vspace{0.5ex}

  \setlength{\tabcolsep}{3.5pt}
  \renewcommand{\arraystretch}{1.1}
  \begin{tabular}{@{}cccccccccccc@{}}
    \toprule
    \textbf{Epochs} & \textbf{Res.} &
    $\boldsymbol{t_{\mathrm{CFD}}}$ & $\boldsymbol{t_{\mathrm{ML}}}$ & $\boldsymbol{t_{\mathrm{up}}}$ &
    $\boldsymbol{n_{\mathrm{switch}}}$ & $\boldsymbol{n_{\mathrm{CFD}}}$ & $\boldsymbol{n_{\mathrm{ML}}}$ &
    \textbf{CFD (s)} & \textbf{ML+CFD (s)} &
    $\boldsymbol{\psi}$ & $\boldsymbol{t_{\mathrm{avg.}}}$ \\
    \midrule
    2  &   5  & 0.75 & 0.029 & 1.31 & 372 & 3713 & 6287 & 7514.9 & 3466.7 & 2.16 & 16.90 \\
    2  &  10  & 0.75 & 0.027 & 1.31 & 330 & 3293 & 6707 & 7587.1 & 3117.9 & 2.43 & 20.32 \\
    2  & 100  & 0.74 & 0.025 & 1.27 & 210 & 2093 & 7907 & 7465.6 & 2029.5 & 3.67 & 37.65 \\
    10 &   5  & 0.74 & 0.028 & 6.59 & 346 & 3453 & 6550 & 7497.8 & 5055.4 & 1.48 & 18.93 \\
    10 &  10  & 0.74 & 0.027 & 6.27 & 307 & 3063 & 6937 & 7473.9 & 4405.5 & 1.69 & 22.59 \\
    10 & 100  & 0.74 & 0.027 & 6.83 & 217 & 2163 & 7837 & 7456.2 & 3307.2 & 2.25 & 46.85 \\
    \bottomrule
  \end{tabular}

  \vspace{1.25ex}
  \textbf{(b) Time-averaged spatial error metrics (Case 3).}
  \vspace{0.5ex}

  \setlength{\tabcolsep}{4pt}
  \renewcommand{\arraystretch}{1.1}
  \begin{tabular}{@{}cccccccccc@{}}
    \toprule
    \textbf{Epochs} & \textbf{Res.} &
    \textbf{L$_2$(T)} & \textbf{MSE(T)} & \textbf{MAE(T)} & \textbf{MaxAE(T)} &
    \textbf{L$_2$(U)} & \textbf{MSE(U)} & \textbf{MAE(U)} & \textbf{MaxAE(U)} \\
    \midrule
    2  &   5  & 2.32e$-$3 & 0.487 & 0.449 & 3.93 & 0.311 & 2.90e$-$4 & 0.013 & 0.096 \\
    2  &  10  & 2.71e$-$3 & 0.666 & 0.540 & 4.39 & 0.367 & 4.07e$-$4 & 0.016 & 0.102 \\
    2  & 100  & 6.55e$-$3 & 4.089 & 1.391 & 8.42 & 0.749 & 1.67e$-$3 & 0.032 & 0.172 \\
    10 &   5  & 2.12e$-$3 & 0.416 & 0.406 & 3.84 & 0.304 & 2.70e$-$4 & 0.013 & 0.084 \\
    10 &  10  & 2.62e$-$3 & 0.670 & 0.516 & 4.30 & 0.368 & 4.10e$-$4 & 0.015 & 0.093 \\
    10 & 100  & 6.21e$-$3 & 4.020 & 1.279 & 8.46 & 0.818 & 2.10e$-$3 & 0.035 & 0.172 \\
    \bottomrule
  \end{tabular}

  \end{threeparttable}
\end{sidewaystable*}


\section{Hardware and software information}
\label{sup_sec:hardsoft_info}
An \texttt{environment.yml} file accompanies the code with the exact package versions. Major components are:
\begin{itemize}
  \item Operating system: Ubuntu 22.04.5 LTS (Linux)
  \item CPU: AMD EPYC 9554 (256-core)
  \item GPU: NVIDIA A100 (40\,GB)
  \item NVIDIA driver: 580.82.9
  \item PyTorch: 2.8.0 + CUDA~12.9
  \item NumPy: 2.2.4
  \item OpenFOAM: v13
  \item SciPy: 1.15.2
  \item Matplotlib: 3.10.0
\end{itemize}

\end{document}